\documentclass[runningheads]{llncs}

\usepackage{eccv}
\usepackage{eccvabbrv}
\usepackage{graphicx}
\usepackage{booktabs}
\usepackage{float}
\usepackage{placeins}
\usepackage{amsmath}
\usepackage{amssymb}
\usepackage{wrapfig}
\usepackage{array}
\usepackage{multirow}

\newcommand{\ours}{\textsc{WildFin}}
\newcommand{\bours}{\textsc{\textbf{WildFin}}}
 \usepackage{hyperref}

\usepackage{orcidlink}

\graphicspath{{assets/}}

\begin{document}

\title{WildFin: An In-the-Wild Dataset for Fish Behavioral Recognition}
 
\author{Abigail G. Grassick\inst{1,*}\orcidlink{0009-0007-2484-0230} \and
Jerome Tze-Hou Hsu\inst{1,*} \and
Ethan Lin\inst{1,*}\orcidlink{0009-0007-6448-690X} \and Ziang Liu \inst{1}\orcidlink{0000-0001-6274-0459} \and Max Whitton \inst{1} \and Madelyn Hair\inst{2}\orcidlink{0009-0009-8771-2871} \and Liam Gutierrez\inst{2} and Haozheng Yu  \inst{1} \orcidlink{0009-0007-6709-3191} \and Kristin Branson\inst{3}\orcidlink{0000-0002-5567-2512} \and Vivek Jayaraman\inst{3}\orcidlink{0000-0003-3680-7378} \and Michael A. Gil\inst{2}\orcidlink{0000-0002-0411-8378} \and \\ Andrew M. Hein\inst{1}\orcidlink{0000-0001-7217-6185} \and Jennifer J. Sun\inst{1}\orcidlink{0000-0002-0906-6589}} 

\authorrunning{Grassick, Tze-Hou Hsu, Lin, et al.}

\institute{Cornell University, 14850 Ithaca, NY \and
University of Colorado Boulder,  80309 Boulder, CO \and HHMI Janelia Research Campus, 20147, Ashburn VA}

\maketitle
\vspace{-2em}
\begin{abstract}
Recent advances in field technology have led to a massive influx of in-the-wild video data for ecological science. The primary bottleneck in leveraging this data is the high cost of expert annotation.
While computer vision offers a potential solution, current models frequently fail when deployed in complex marine environments. To characterize these failures, we introduce \ours, a novel benchmark for fish behavior recognition collected and annotated by ecologists.\ours{} spans two critical real-world paradigms: stationary cameras monitoring groups of fish and dynamic divers following individual subjects. The dataset represents a massive curation effort, involving 1,350 hours of fieldwork and 600 hours of expert annotation to produce 9 hours of behavioral data with over 2 million frame-by-frame labels.
We benchmark modern vision foundation models and quantify tradeoffs between static and spatiotemporal architectures, revealing the substantial gap that remains between current model capabilities and the demands of real-world underwater behavioral analysis. Project website: \url{https://team-wildfin.github.io/}.

\begingroup
\renewcommand{\thefootnote}{*}
\begin{NoHyper}
\footnotetext{Equal contribution. Contact: agg75@cornell.edu.}
\end{NoHyper}

\endgroup

\end{abstract}

\vspace{-3em}

\section{Introduction}
\label{sec:intro}
\vspace{-0.8em}

\begin{figure*}[t]
    \centering
    \includegraphics[width=\textwidth]{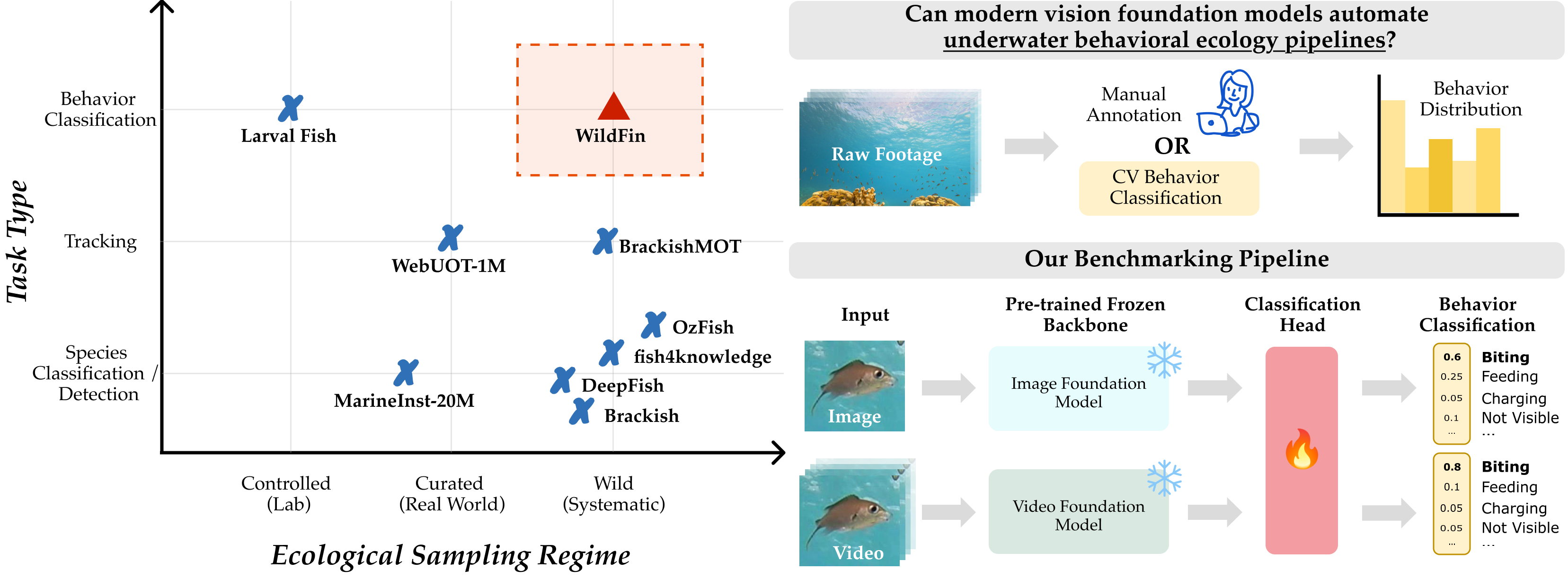}
    \caption{\scriptsize \textbf{\ours: Dataset Overview.} \ours{}  fills a field-wide gap in publicly available underwater behavioral video datasets. Datasets displayed in our chart are categorical, and marker placement within each section is arbitrary. The videos featured in \ours{}  were collected via a field ecology pipeline and, once processed, are used to derive biological insights from the distributions of observed behaviors. Our benchmarking pipeline passes input images or video clips through a pre-trained, frozen backbone to obtain behavioral classification outputs.}
    \label{fig:overview}
    \vspace{-2em}
\end{figure*}

The behavior of marine organisms is essential in governing the dynamics of the health and maintenance of marine ecosystems \cite{gil2020fast,ling2015global,urmy2021fear,filbee2014sea}. 
Observing these behaviors in natural settings provides invaluable context for  ecological science~\cite{gil2020fast,harrold1985food}, conservation \cite{gil2020fast, harrold1985food} as well as neuroscience research \cite{hein2018conserved, fahimipour2023wild}, where many marine species serve as pivotal models~\cite{fetcho2012crystal,hochner2012embodied,kang2023sensory,zada2024development}. While field-deployable video camera systems have revolutionized the scale at which behavioral data can be collected \cite{belcher2023demystifying, hughey2018challenges}, this capacity has introduced a critical data-processing bottleneck: the sheer volume of footage far exceeds the capacity for human analysis. 
Manual post-processing is increasingly intractable; for example the two subsets of data presented in \ours, CoralCam and FishFollow (Fig.~\ref{fig:overview}), required 280 and 320 expert-hours of annotation respectively. 

While automated methods for object detection and classification in still marine images have been studied~\cite{belcher2023demystifying}, powered by datasets such as~\cite{katija2022fathomnet,saleh2020realistic}, extending these methods to dynamic, video-based behavior remains a significant frontier. Robust automated analysis of in-situ fish behavior poses unique difficulties not fully captured by existing human action recognition benchmarks \cite{kay2017kinetics,goyal2017something} or other animal behavior benchmarks, which typically focus on terrestrial \cite{gabeff2025mammalpsmultiviewvideobehavior, kholiavchenko2024kabr}  or laboratory \cite{CalMS21, elsayed2011flyvsfly} settings. The challenges to this problem include: fast-moving subjects, subtle behaviors (e.g., grazing vs. aggression), rare behaviors, and severely obstructive visual conditions unique to underwater imagery, including dynamic illumination, back-scatter, and variable water clarity 
\cite{rout2024underwater, kapoor2025graph, nissar2024nicasu}. 
The scarcity of public, expert-annotated benchmarks capturing these complex dynamics has curtailed the development and  evaluation of automated approaches.

To address this gap, we introduce \ours, a novel video benchmark built from real-world ecological field data, rather than curated specifically for computer vision (Fig.~\ref{fig:overview}). The included footage was collected by field biologists as part of ongoing coral reef research and, until now, remains largely unanalyzed due to the extensive cost of manual annotation. \ours \space has not been previously published or used outside of this research context.

The dataset itself consists of approximately 9.2 hours of high-resolution (1080p and 4K) video capturing freely-behaving fish, with over 2 million frame-by-frame expert-annotated labels, organized into two distinct subsets reflecting real-world collection scenarios:
\vspace{-0.5em}
\begin{itemize}
    \item CoralCam: 1.2 hours (213 selected tracks of individuals) of stationary, multi-agent video from 12 distinct reef habitats. This subset provides a unique multi-stage resource, including the original object detections and expert-verified tracks used to generate behavioral labels for the annotated behaviors.
    \item FishFollow: 8 hours (81 videos) of dynamic, focal-follow video tracking single fish, mimicking the egocentric footage captured by divers or citizen-scientists in-the-wild. It includes fine-grained behavioral categories spanning social interactions and solitary actions.
\end{itemize}
\vspace{-0.5em}

\ours{} is representative of existing captured data found across many ecological settings. 
While our dataset benchmarks behavior in underwater coral reef systems, the shortcomings of current models apply broadly to other ecological fields as well.
This is because, unlike datasets curated specifically for computer vision, \ours{} was collected using standard ecological field protocols; thus, it captures the complexities (e.g. dynamic lighting, occlusions, and varied perspectives) that current models must overcome to be useful for scientific discovery. 

\begin{figure*}[t]
    \centering
    \includegraphics[width=1.0\linewidth]{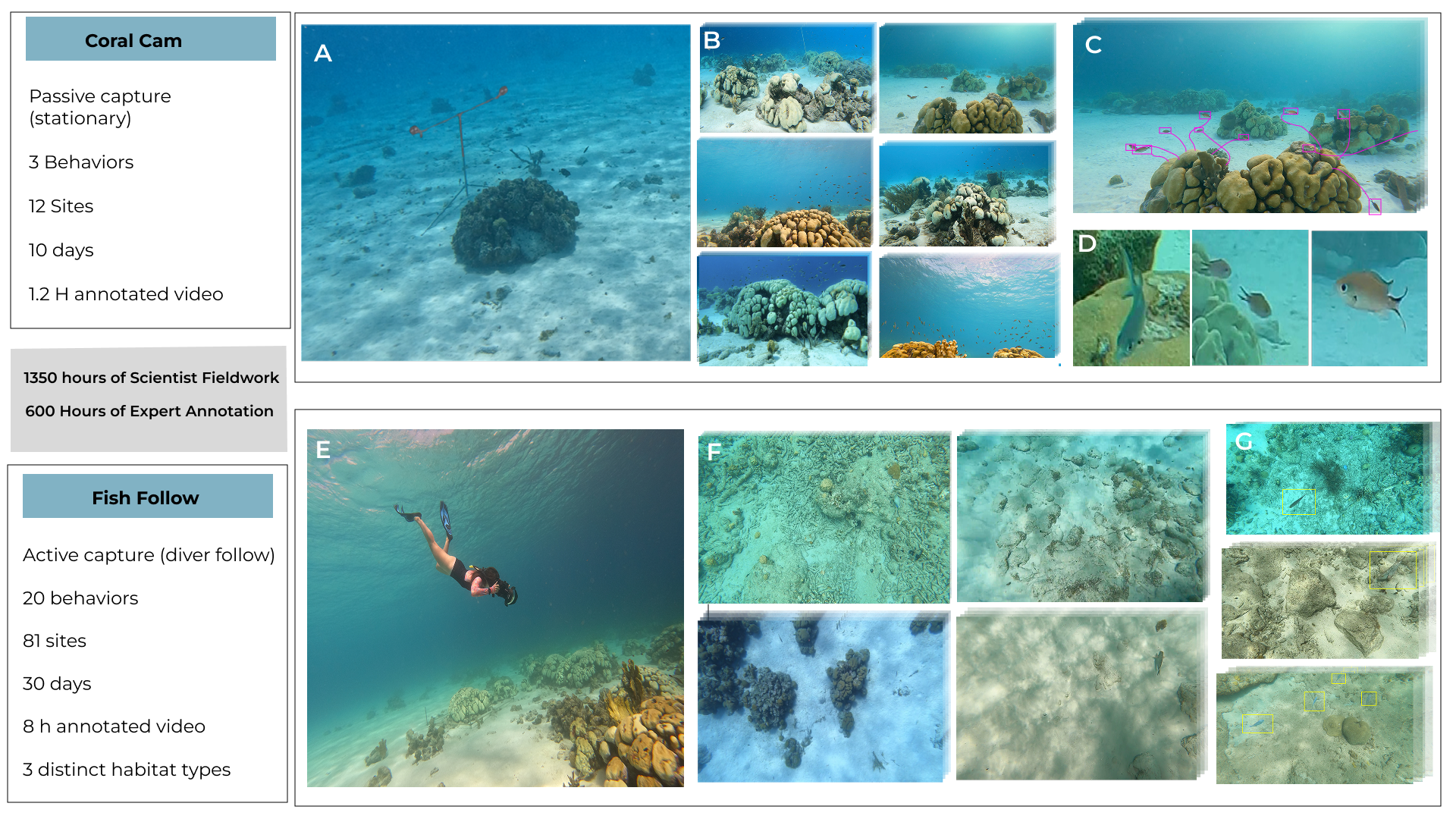}
    \vspace{-1em}
    \caption{\scriptsize \textbf{\bours{} data types and process: } \ours{}  comprises two subsets: CoralCam (stationary) and FishFollow (dynamic). (A) Tripod setup used for CoralCam data collection. (B) Six representative example sites illustrating the range of scenes found in this subset. (C) Videos are processed through an object detection and multi-object tracking (MOT) pipeline, (D) an annotator categorizes the behavior of each individual on a frame-by-frame basis; behaviors shown here include feeding, aggression, and not feeding. (E) The dynamic capture style of FishFollow, in which a diver tracks a single, free-swimming fish. (F) Sample frames from FishFollow videos; note that backgrounds and habitats vary widely even within a single video. (G) Annotators perform multi-frame state behavior annotations across 23 behavior classes; examples shown here include solo foraging, sand rubbing, and social foraging.
 }
    \label{fig:dataset}
    \vspace{-2.5em}
\end{figure*}

Importantly, \ours{} pools data from two distinct, but related, data collection and annotation paradigms (Fig.~\ref{fig:dataset}). The heterogeneity of this dataset is not a weakness, rather the primary strength of its contribution; \ours{} is representative of the limitations of ecological datasets: they are messy, taken under wild field constraints, and often undergo variety in collection and annotation protocols year to year based on these constraints. We argue that benchmarking capturing this kind of real-world variability are valuable in that they better prepare computer vision methods for deployment in genuine ecological monitoring contexts, rather than the more controlled conditions of existing datasets. 

\vspace{5pt}
We summarize our main contributions as follows:
\vspace{-5pt}
\begin{itemize}
    \item We introduce \ours, a video benchmark uniquely derived from ongoing ecological research spanning two distinct real world capture scenarios. 
   
    \item We establish baselines using leading vision foundation models, quantifying the trade-offs between static and spatio-temporal architectures. Our results highlight shortcomings in the performance of current models and demonstrate the necessity of different strategies (e.g. lightweight pooling vs. fine-tuning, class imbalance mitigation) to navigate the unique challenges of underwater behavioral analysis.
\end{itemize}
\vspace{-2em}
\section{Related Work}
\vspace{-1em}

\label{sec:related}

\paragraph{\textbf{Video analysis and vision foundation models:}}
Recent work in video analysis has increasingly focused on the application of vision-based foundation models instead of using task-specific architectures \cite{yuan2024videogluevideogeneralunderstanding,zhao2025videoprismfoundationalvisualencoder,videomae}. These large models are typically pre-trained on massive datasets using self-supervised learning to acquire transferable representations that generalize across domains. 

Modern video foundation models incorporate temporal modeling through attention mechanisms or temporal masking and achieve strong performance on tasks like action recognition~\cite{TimeSformer,videomae,videomaev2,vivit,assran2025vjepa2selfsupervisedvideo}. Although trained on static images, models like CLIP~\cite{clip} and DINO~\cite{dinov2,siméoni2025dinov3} are frequently adapted for video tasks, often serving as powerful and efficient baselines. In this work, we benchmark a diverse range of foundation models including DINOv3 ~\cite{siméoni2025dinov3}, VideoMAE ~\cite{videomaev2}, and V-JEPA 2 ~\cite{assran2025vjepa2selfsupervisedvideo}, on the task of fish behavior classification in in-situ underwater video. 

\paragraph{\textbf{Methods and Datasets for Underwater Imagery:}}
Despite growing interest in underwater vision, most publicly available datasets in this domain consist of still images, rather than video \cite{belcher2023demystifying}. Notable still-image datasets include the annotated fish classification dataset by Saleh et al.\cite{saleh2020realistic}, the broad taxonomic coverage of the \textit{FathomNet} database\cite{katija2022fathomnet}, and the \textit{MarineInst} dataset~\cite{ziqiang2024marineinst}, which combines field-collected and web-sourced imagery. Most existing methods for analyzing underwater imagery in ecological contexts have also focused on still images, including detecting, classifying, and segmenting marine animals (e.g., \cite{laradji2021weakly,jalal2020fish,knausgaard2022temperate,katija2022fathomnet,kyathanahally2022ensembles,talluri2024ensemble, beery2019efficient}). 

A smaller number of underwater video datasets exist, including those with object-level annotations such as bounding boxes or segmentation masks (\cite{ditria2021annotated,Fisher2016,ozfish2019dataset, Pedersen_2019_CVPR_Workshops}). However, these datasets lack comprehensive behavior annotations, limiting their utility for fine-grained ecological analyses. Few studies have explored video-based computer-vision analysis for underwater imagery. These approaches apply object detection and tracking, followed by classification based on the animals' trajectories \cite{hein2018conserved,fahimipour2023wild, pedersen2023brackishmotbrackishmultiobjecttracking}. 
In summary, there is a clear gap in publicly available datasets that provide ecologically relevant underwater video with expert behavioral annotations. \ours{} is introduced to address this gap and facilitate research at the intersection of marine ecology and video understanding.

\paragraph{\textbf{Computer vision for animals:}}
Computer vision for animals is an increasingly important area at the intersection of computer science and biology, enabling scalable analysis of in-situ animal behavior, movement, and species interactions~\cite{anderson2014toward,sun2024video,pereira2020quantifying}. This research supports advances across neuroscience, behavioral analysis, and ecology~\cite{CalMS21,gabeff2025mammalpsmultiviewvideobehavior}. Large-scale annotated datasets have played a central role in driving model development for these domains. This includes studies in laboratory settings (e.g. mice~\cite{CalMS21}, flies~\cite{elsayed2011flyvsfly}, rats~\cite{pairr24m}, and larval fish ~\cite{bar2022analysis}), zoo environments (e.g. chimps~\cite{ma2023chimpact}), in natural land habitats (e.g. animals in the alps~\cite{gabeff2025mammalpsmultiviewvideobehavior}, and Kenya~\cite{kholiavchenko2024kabr}) as well as underwater (e.g. fish segmentation~\cite{saleh2020realistic} or detection~\cite{ozfish2019dataset, Pedersen_2019_CVPR_Workshops}).

However, there are far fewer large-scale annotated datasets of animals in-the-wild. Compared to controlled lab environments, collecting behavioral data in the wild introduces substantial challenges: highly variable lighting, unrestricted camera viewpoints, and visually cluttered or occluded backgrounds. MammAlps \cite{gabeff2025mammalpsmultiviewvideobehavior} illustrates some of these difficulties in terrestrial settings; however, in underwater environments, these challenges are even more severe due to dynamic illumination, varying turbidity, and poor contrast. Furthermore, fish themselves are non-standard subjects in computer vision—lacking consistent anatomical landmarks, often displaying reflective or translucent bodies, and exhibiting a wide range of morphologies and movement dynamics. Variability in the time scales of behavior introduces more challenges, as fish display actions ranging from extended foraging to sub-second escape maneuvers \cite{hein2018conserved}. These compound difficulties, combined with a scarcity of behavior-labeled underwater video datasets, have limited progress in automated fish behavior understanding.

Another core challenge is that animal behavior is highly imbalanced: common behaviors dominate, while rare but ecologically important actions occur infrequently. This long-tailed distribution, seen in datasets like CalMS21 \cite{CalMS21}, Fly vs Fly \cite{elsayed2011flyvsfly}, and MammAlps \cite{gabeff2025mammalpsmultiviewvideobehavior}, poses challenges for obtaining many training examples of rare behaviors and learning reliable models. Techniques such as focal loss \cite{lin2018focallossdenseobject}, class weighting \cite{xu2020classweightedclassificationtradeoffsrobust}, and balanced sampling strategies \cite{chen2024imbalanced} have been used to address this. These class imbalances are complicated in ecological research by the fact that data collection is often expensive and labor intensive, and manually annotating more data is not a scalable solution. Thus, methods for ecological computer vision must be able to overcome the challenges of long tailed distributions in order to be effective for the field. In our work, we benchmark balanced batch sampling and focal loss techniques and highlight the need for more data-efficient, imbalance-aware methods, especially in ecological settings where rare behaviors carry high scientific value.

\vspace{-1.25em}
\section{Dataset}
\vspace{-1em}
\label{sec:dataset}

The \ours{} benchmark is composed of two annotated subsets, \textbf{CoralCam} (stationary multi-agent reef recordings) and \textbf{FishFollow} (mobile focal-follow videos), capturing wild
marine fish behavior in real-world ecological settings. Together they contain 9.2 hours (2,058,892 frames) of expert-annotated video, spanning 23 behaviors across both subsets (Fig.~\ref{fig:dataset}). All annotations and pipeline artifacts (detections weights, annotations, tracks, and behavioral annotations) are available on our project website: \url{https://team-wildfin.github.io/}. 
The benchmark reflects a real-world ecological video-analysis pipeline andp resents substantial spatiotemporal challenges for modern video foundation models. \ours{}'s two subsets arise from the primary paradigms of underwater visual data collection and are characteristic of the many ecologist-curated datasets that predate the use of computer vision as an analytical tool. This section details our dataset, starting with our general curation and annotation philosophy (\S\ref{sec:annotation}), our two data-collection paradigms (\S\ref{sec:paradigms}), and detailed descriptions of our annotated subsets, CoralCam (\S\ref{sec:coralcam}) and FishFollow (\S\ref{sec:fishfollow}).
\vspace{-1em}
\subsection{Data Curation and Annotation}

\label{sec:annotation}

The primary task defined by \ours{} is \textbf{frame-level, multi-label behavior
classification}: given a video clip and a tracked individual, assign a
behavioral label to every frame.

Secondary tasks, such as object detection and multi-object tracking, form the upstream pipeline that produces the tracked individuals on which classification is evaluated, and we release annotations and benchmarks for each. Evaluation is therefore performed at the finest temporal granularity the annotations support: individual frames, aligned to expert-verified tracks.

\begin{wrapfigure}{r}{0.48\linewidth}
    \centering
    \vspace{-2.5em}
    \includegraphics[width=\linewidth]{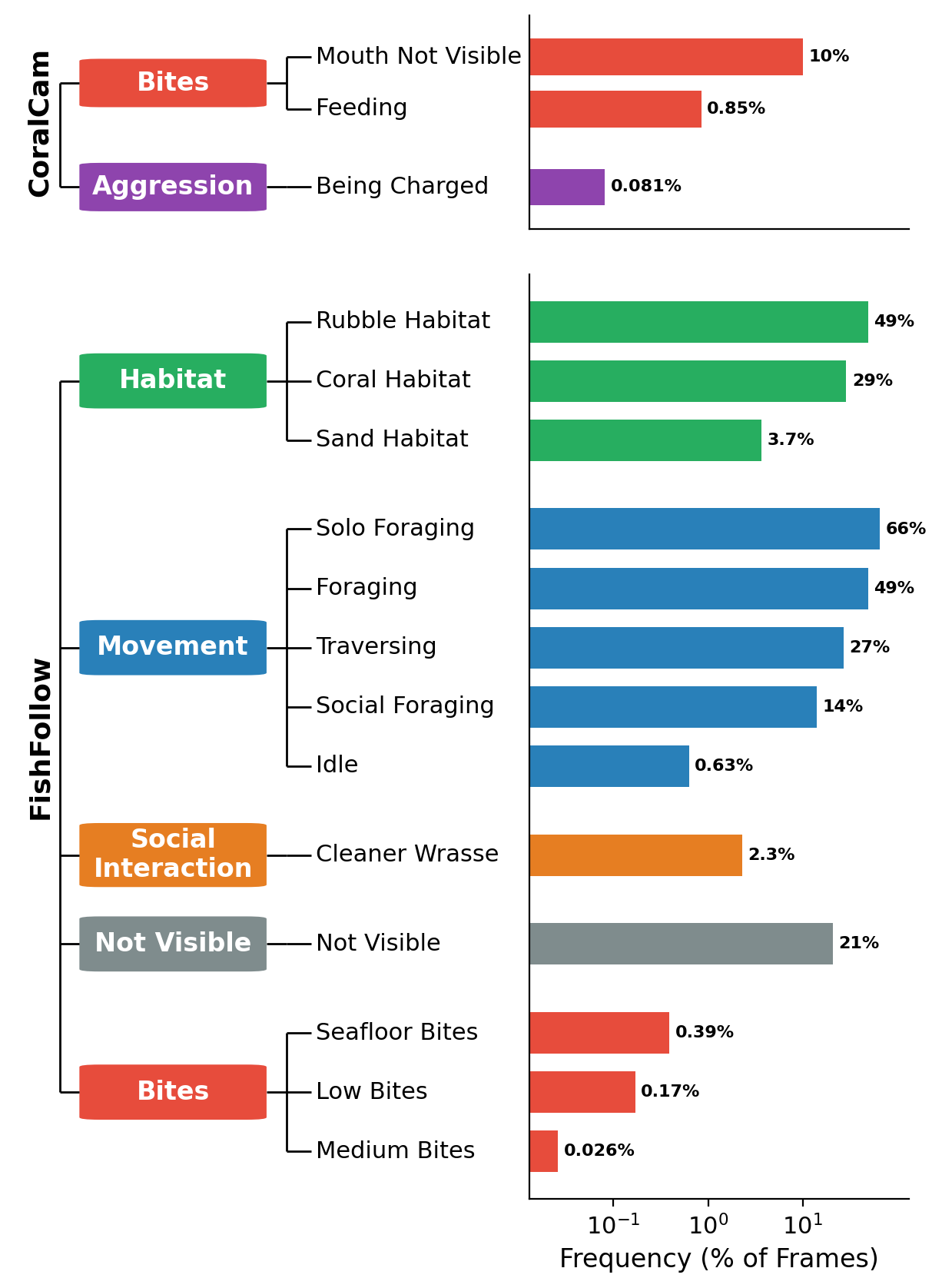}
    \vspace{-2.5em}
    \caption{\scriptsize \textbf{Behavioral distributions:} Behavioral distributions found in both CoralCam and FishFollow. CoralCam behaviors can be split into bite and aggression classifications, while FishFollow is composed of three distinct behavioral types and one habitat interaction class. Percentages indicate the fraction of total video frames annotated with each behavior class; because a large proportion of frames are unannotated (null/background), percentages do not sum to 100\% and instead reflect each class's prevalence relative to the full dataset. Furthermore, a single frame may contain several behavioral annotations. For example, an individual may be solo foraging and taking bites at the seafloor at the same time, reflecting the multilabel nature of this dataset.}
    \label{fig:distributions}
    \vspace{-2.7em}
\end{wrapfigure}
The ecological behaviors captured in \ours{} map directly onto a set of
computer-vision challenges that challenge modern spatiotemporal models. Many behaviors, such as sustained bouts of feeding or prolonged displays of aggression, unfold over seconds to minutes, demanding \emph{long temporal
dependencies} that short clip models cannot resolve. Discriminating between superficially similar actions (e.g., a feeding strike vs. a charge) requires sensitivity to \emph{fine-grained motion cues} spanning only a handful of frames. In CoralCam, schools of tens to hundreds of individuals produce \emph{severe multi-agent occlusion}, where focal animals are frequently partially or fully obscured by conspecifics. In FishFollow, a diver-mounted camera introduces continuous \emph{ego-motion}, rapid \emph{viewpoint changes}, and \emph{dynamic backgrounds} as the focal fish traverses multiple habitat zones. Taken together, these properties make \ours{} a rigorous testbed for spatiotemporal representation learning under ecologically realistic conditions.

\textbf{Scientifically-Grounded Behaviors.} Every behavior in \ours{} is drawn from formal ethograms established by coral reef ecologists \cite{branconi2019comparison, matthews2015temperature, sopinka2009liver}; specifically, these data are being used in active unpublished research projects. A complete ethogram with per-behavior descriptions is provided in the appendix. 

\textbf{Naturalistic / Wild Environments.}
All videos were recorded in wild reef environments in Cura\c{c}ao by field biologists, with no management of fish behavior or habitat. Unlike many behavioral datasets collected in controlled laboratory or semi-wild settings, \ours{} displays authentic social and solitary
interactions across a diverse reef ecosystem.
\vspace{-1\baselineskip}
\subsection{Data-Collection Paradigms}
\label{sec:paradigms}

The benchmark intentionally spans the two primary paradigms of underwater ecological video collection, each of which introduces a distinct set of
computer-vision challenges.

\textbf{Passive Sampling : Stationary Underwater Video (CoralCam).}

Fixed, tripod-mounted cameras are deployed on the reef and left to record continuously for many hours at a time. This passive approach is common in ecological monitoring because it
requires minimal observer presence and can be replicated across many sites simultaneously.
Frames are populated by dense schools of small fish against highly textured coral backgrounds, suspended particulate, variable lighting, and visual noise from wave action producing severe \emph{multi-agent occlusion}, \emph{small-object detection} at low apparent resolution, and substantial \emph{background clutter} from benthic structures and other light caustics associated with underwater scenes.

\textbf{Active Sampling : Mobile Focal-Follow Video (FishFollow).}
A diver follows a single individual with a handheld camera as it moves throughout its habitat, capturing interactions with its surroundings and with other species within the landscape. This paradigm is a standard approach for studying individual behavioral ecology, and is found to have minimal disturbance to fish behavior \cite{branconi2019comparison}.

The moving-camera design introduces persistent \emph{ego-motion},
\emph{large viewpoint changes}, \emph{scale variation} as the focal fish
approaches or recedes, and \emph{dynamic backgrounds} as the diver-fish pair traverses
structurally distinct habitat patches.

\subsection{Annotated Subsets}
\label{sec:subsets}

\subsubsection{CoralCam}
\label{sec:coralcam}
\vspace{-5 pt}
\paragraph{\textbf{Collection and Context:}}
CoralCam provides fixed-site observations of schools of
\textit{Azurina multilineata} (brown chromis) across 12 distinct reef sites
in Cura\c{c}ao.
Video was captured at depths of 3--5\,m using stationary, tripod-mounted
GoPro Hero 9 cameras at 1080p@60\,fps, yielding 1.2 hours (258,998 frames)
of footage.
Fixed-site recordings provide consistent, repeatable coverage of defined
spatial areas, enabling the study of group behavior, predator–prey
interactions, and site-specific dynamics.
\vspace{-8 pt}
\paragraph{\textbf{CV Pipeline and Annotation:}}
Each frame typically contains tens to hundreds of individuals, making
direct full-frame annotation intractable.
We therefore follow a three-stage pipeline used by ecologists to annotate
ecological video datasets.
\begin{itemize}
    \item \textit{Object Detection.} Ecologists hand-labeled bounding boxes
    for three species of interest (21,116 expert-labeled boxes).
    These annotations were used to train a YOLOv8 detector~\cite{10533619},
    which achieved AP@0.5 of 74.5 on a held-out test set.
    We release these annotations in COCO~\cite{lin2014microsoft} format as an
    in-situ fish detection and classification benchmark.

    \item \textit{Multi-Object Tracking.} The trained detector was deployed
  across sites and per-frame detections were linked into tracks using the
    BoTSort algorithm~\cite{aharon2022botsortrobustassociationsmultipedestrian}.

    \item \textit{Expert Verification \& Behavioral Annotation.} 
    A human expert selected and verified 213 tracks from the analyzed sites,
    then applied frame-by-frame behavioral labels for 3 behaviors:
    \textit{feeding}(mouth open and mouth closed), and \textit{charging}. 
    \textit{Mouth visibility} was additionally encoded to account for partial
    occlusion and individuals facing away from the camera, resulting in the annotators' inability to classify the frame. 
    Inter-annotator F1 on a held-out subset was 0.74; further verification
    metrics and annotator-bias analyses are in the appendix.
\end{itemize}
\vspace{-5pt}
The primary task for CoralCam is \textbf{frame-level behavior classification
on tracked individuals in multi-agent stationary scenes}: given a
track and its corresponding frame crops, predict the behavioral label for
each frame.
\vspace{-5 pt}
\paragraph{\textbf{Limitations:}} CoralCam's annotation pipeline is decomposed into several dependent stages, so errors introduced early propagate through the remainder of the pipeline. The detector is the first such source: its AP@0.5 of 74.5 upper-bounds the quality of every downstream stage. This aggregate figure conflates two distinct error regimes. For foreground individuals, the only ones for which behavioral annotation is feasible, since the relevant signals are subtle and require close, un-occluded views, the dominant failure is species confusion rather than missed detection, and the resulting tracks remain clean and continuous. A confusion matrix outlining per-class performance can be found in the appendix. Missed detections instead concentrate on small, distant individuals, which are excluded from the behavioral set entirely. Both regimes bias the data the pipeline produces: the detection benchmark inherits noisy species labels, and the behavioral benchmark is implicitly restricted to near-field fish.
\\\\
Tracking introduces a second, harder-to-quantify limitation: constructing ground-truth tracks for these videos is not feasible, so in its current form the dataset offers no reliable way to evaluate tracking quality. Exhaustively tracking every individual across the full frame is ruled out by the tens to hundreds of fish present at any moment. Furthermore, exhaustively ground truthing the individuals within a bounded spatial region of the frame does not recover true performance either, as fish continually enter, leave, and occlude one another across the region boundary. Therefore, a region-restricted evaluation neither captures the tracker's behavior on the individuals that cross it nor generalizes to the full scene. Short tracks appear comparatively reliable under this partial view, but capturing a behavior of interest requires following an individual long enough for that behavior to manifest, so annotators instead selected a subset of longer, manually verified tracks (213 in total). This selection has direct downstream consequences: it favors longer tracks with fewer occlusions and more standardized backgrounds, leaves identity switches during occlusion unquantified, and under-represents fast or erratic behaviors such as charging (precisely the behaviors most likely to fragment a track) thereby biasing the behavioral set toward calmer, more easily followed individuals.

\subsubsection{FishFollow}
\label{sec:fishfollow}
\vspace{-1.5 pt}
\paragraph{\textbf{Collection and Context:}}
FishFollow offers dynamic, focal-follow recordings of
\textit{Labriformes} (parrotfish) and \textit{Acanthuridae} (surgeonfish).
The subset comprises 81 videos totaling 8 hours (1,799,894 frames) of footage,
captured with GoPro Hero 9 cameras at 4K@60\,fps.
Each video follows a single focal fish as it moves through varying habitats
and social states, mirroring how an expert or citizen scientist would record
behavioral data in the field.
This design facilitates future integration of crowd-sourced footage from
citizen scientists.
\vspace{-8pt}
\paragraph{\textbf{CV Pipeline and Annotation:}}
Because each video focuses on a single focal animal, frame-level annotations
could be applied directly without an upstream tracking pipeline.
A team of trained annotators labeled 20 behaviors spanning social, solitary,
feeding, and inter-species interactions.
To ensure label quality, 35\% of the videos were independently annotated by a
second expert, yielding a high inter-annotator agreement (average F1 of 0.78
with a 0.25\,s tolerance; F1 increases to 0.93 when tolerance is increased to 0.5s).

The primary task for FishFollow is \textbf{frame-level behavior
classification under dynamic viewpoint and habitat transitions}: given the
full focal-follow video, predict the behavioral label for each frame as the
focal individual and camera move continuously through the scene.

\vspace{-8 pt}
\paragraph{\textbf{Limitations:}} FishFollow uses a deliberately minimal pipeline, using direct frame-level annotation with no upstream tracking. This pipeline shifts the burden onto a single unresolved problem: the focal individual is never spatially localized. Each video establishes the focal fish through the diver's initial pointing gesture, but many conspecifics may share the frame, and no bounding-box or segmentation annotation ties the frame's behavior label to that individual thereafter. This leaves the label without an explicit target, and a model cannot reliably determine which fish the label refers to. Thus, models may instead learn to associate a behavior with its presence anywhere in the frame rather than in the focal animal. Identity across the video rests on an implicit camera-centering prior that is neither annotated nor dependable through occlusions, habitat transitions, and the inevitable variability introduced by currents and wave action. The absence of localization annotations is therefore the principal factor limiting model performance on this dataset. A second limitation is visible in the annotation agreement itself. Average inter-annotator F1 is high (0.78 at a 0.25 s tolerance), but rises to 0.93 when the tolerance is widened to 0.5 s, indicating that most disagreement is concentrated at behavioral boundaries rather than in the behaviors assigned: annotators agree on what occurs but not on its precise onset and offset. Because FishFollow's task is defined under continuous viewpoint and habitat transitions, this boundary uncertainty falls disproportionately on the frames the task most depends on, and frame-exact evaluation will penalize models for the same transitions humans annotate inconsistently.
\vspace{-1em}
\section{Benchmark Design and Protocol}
\vspace{-1em}
\label{sec:eval_method}

To evaluate the capabilities of recent vision foundation models on the task of in-situ fish behavior classification, we benchmark a diverse set of backbones, fine-tuning strategies, and class-imbalance mitigation techniques on \ours{}. Following prior works in foundation model adaptation~\cite{yuan2024videogluevideogeneralunderstanding, zhao2025videoprismfoundationalvisualencoder, hasson2025scividcrossdomainevaluationvideo, sun2024video}, we adopt the common paradigm of using a frozen pre-trained backbone while fine-tuning a lightweight pooling and classification head on top. We also explore methods to address the strong dataset imbalance inherent in ecological behavior data. In sections 4.1 and 4.2, we discuss our experimental setup, metrics and design of experimental variables on the main fish behavior classification subsets of \ours{}, CoralCam and FishFollow. In section 4.3, we discuss how we benchmark and evaluate different models on the object detection dataset.
\vspace{-0.5\baselineskip}
\subsection{Experimental Setup}
\textbf{Data Splits:} We split the videos in both CoralCam and FishFollow into approximately 70\% train and 30\% test, ensuring that videos across splits are captured from different sites/dates.

\textbf{Preprocessing:} Our models fall into two categories. For video-based models (VideoMAE, V-JEPA2), we use 16 frame windows, sampling every frame. For image-based models (DINOv3, ResNet50), we process single frames. For the CoralCam subset, the model receives cropped videos or frames of single fish. For the FishFollow subset, the model processes full-size videos. In both cases, the pipeline downsizes frames to $224 \times 224$ px to fit standard model input. 

\textbf{Metrics:} We evaluate model performance using macro-averaged F1, precision, and recall. A key challenge in this task is the ambiguity of event boundaries. A tolerance of 0 frames, which requires an exact frame match, may not reflect the real-world use case. Following input from ecologists, we report our primary results using a tolerance of 7 frames, which corresponds to a temporal tolerance of approximately $\pm 0.1$ seconds for videos recorded at 60~FPS.
This considers a prediction correct if it falls within a 7-frame window of a true positive, reflecting the fact that point-based behaviors manifest over short temporal intervals.
\vspace{-0.5\baselineskip}
\subsection{Experimental Variables}

\textbf{Vision Backbones:} We evaluate backbones from distinct training paradigms and architectures. Our selection is designed to test two key dimensions for analyzing our underwater ecology data: (1) the value of modern vision foundation models against classical CNNs (like ResNet50) that are commonly used in scientific pipelines, and (2) the performance trade-offs between distinct processing strategies: static frame-by-frame versus spatio-temporal.
\\ \textit{Vision Foundation Models.} This group represents the modern, state-of-the-art approach, including image-based (DINOv3-B, DINOv3-L) and video-based (V-JEPA2-L, VideoMAE-B, \&VideoMAE-L) models. 
DINOv3 is a state-of-the-art, transformer-based image model trained using a self-supervised objective \cite{siméoni2025dinov3}.
The video models are transformer-based video models that explicitly model the temporal dimension. Both models are  trained using self-supervised objectives that predict masked spatio-temporal regions.
We chose these two models to represent distinct, practical points in the design space. VideoMAE-B~\cite{videomae} serves as a widely-adopted and efficient (ViT-Base) baseline. In contrast, V-JEPA2-L~\cite{assran2025vjepa2selfsupervisedvideo} represents a more recent, state-of-the-art (ViT-Large) model.
\\ \textit{Classic Supervised Baseline.} To provide a reference point, we include a ResNet50 ~\cite{he2016deep} model pretrained on ImageNet~\cite{deng2009imagenet}. This allows us to directly contrast the performance of a commonly used CNN against the modern, self-supervised ViT-based models. Moreover, the lighter weight of ResNet50 also allows us to perform full fine tuning, providing a strong baseline for what is achievable on practical, resource-constrained pipelines.
\\ \textit{Static appearance vs. spatio-temporal dynamics.} To evaluate the performance of a static, frame-by-frame processing strategy, we use our image-native models (DINOv3-B, DINOv3-L, ResNet50). This approach evaluates these models in their most direct application, omitting explicit temporal information. This baseline allows us to quantify which behaviors in \ours{} are solvable using only the static appearance cues captured by these models and which behaviors require temporal information to be identified.

We also evaluate standard, end-to-end video foundation models (V-JEPA2-L, VideoMAE-B, VideoMAE-L). These models process 16-frame clips and are explicitly pre-trained on and designed to model spatio-temporal dynamics. By comparing the performance of these two distinct baselines, we can identify performance trade-offs and reveal which behaviors in \ours{} are better suited to one strategy over the other.
\\ \textit{Addressing Class Imbalance:} Natural behavioral datasets exhibit significant class imbalance ~\cite{CalMS21, mabe22, elsayed2011flyvsfly}, a core challenge in \ours{} as well. On \ours{}, naive random sampling approach generally fails, yielding near-zero performance on rare behaviors (see appendix). To establish a robust and practical baseline, we evaluate two of the most foundational and widely-adopted techniques for this problem:
\begin{itemize}
    \item Balanced Sampling \cite{chen2024imbalanced}: Giving equal sampling probability across positive behavior classes and the null class.
    \item Focal Loss \cite{lin2018focallossdenseobject}: A loss function that down-weights the contribution of common, easy negatives and up-weights rarer, difficult positives.
\end{itemize}

Our goal is to quantify the severity of the imbalance and provide a strong, accessible baseline for practical use. We posit that \ours{} is a challenging testbed where approaches such as retrieval or metric learning can be developed and benchmarked in future work.
\\ \textit{Adaptation Approach:} With the backbone frozen, we must aggregate the token embeddings (for ViT models) or the final feature map (for ResNet) before passing them to the classifier \cite{hasson2025scividcrossdomainevaluationvideo}. We compare two strategies for this lightweight adaptation head:
\begin{itemize}
    \item Mean Pooling: A simple, non-parametric head that uniformly averages all token embeddings into a single feature vector.
    \item Cross-Attention Pooling: A lightweight, learnable head that uses a single query to adaptively weight and aggregate the token embeddings, allowing the model to focus on the most salient features for the task. 
\end{itemize}

The resulting feature vector from either head is then passed to a two-layer MLP classification head.
\vspace{-0.5\baselineskip}
\subsection{Object Detection Benchmark}

The object detection dataset is provided in COCO format with 70\% train, 20\% val, and 10\% test splits. We benchmark a diverse set of widely used detection models covering both convolutional and transformer-based architectures for the task of fish detection and classification in the wild. Specifically, we evaluate Faster R-CNN, YOLO, and RT-DETR, representing common convolutional and transformer-based approaches. 

Final performance is reported on the test split, and more details will be available in the appendix.

We evaluate using standard COCO metrics, reporting Average Precision (AP) at IoU thresholds of 0.5, 0.75, and mAP@[.5:.95]—the mean AP across IoUs from 0.50 to 0.95 in 0.05 increments.
\vspace{-1em}
\section{Benchmark Analysis} 
\vspace{-1em}
\label{sec:results}

We evaluated all combinations of backbones, pooling strategies, and class imbalance techniques on both CoralCam and FishFollow. Comprehensive results are provided in the appendix. For clarity in this section, we report comparisons by fixing a design choice (e.g., backbone or pooling method) and selecting the configuration that achieves the highest F1 score.
\begin{table*}[t]
\centering
\scriptsize
\setlength{\tabcolsep}{2pt}
\renewcommand{\arraystretch}{1.2}

\begin{tabular}{lccc|cccccc}
\toprule

& \multicolumn{3}{c|}{\textbf{CoralCam}} 
& \multicolumn{6}{c}{\textbf{FishFollow}} \\

\cmidrule(lr){2-4} \cmidrule(lr){5-10}

\textbf{Backbone} & \textbf{Overall} & \textbf{Aggression} & \textbf{Biting}
& \textbf{Overall} & \textbf{Habitat} & \textbf{Movement} & \textbf{Bites} & \textbf{Social} & \textbf{Not Vis.} \\

\midrule

DINOv3-B   & 0.309 & 0.000 & 0.464 & 0.378 & 0.582 & 0.465 & 0.108 & 0.055 & 0.457 \\
DINOv3-L   & 0.329 & 0.007 & 0.490 & 0.380 & \textbf{0.594} & 0.458 & 0.123 & 0.036 & 0.458 \\
ResNet50   & 0.286 & 0.000 & 0.429 & 0.361 & 0.571 & 0.460 & 0.082 & 0.023 & 0.405 \\
VideoMAE-B & 0.362 & 0.311 & 0.388 & 0.370 & 0.552 & 0.454 & 0.105 & 0.085 & \textbf{0.486} \\
VideoMAE-L & \textbf{0.450} & \textbf{0.332} & \textbf{0.510} & \textbf{0.387} & 0.563 & 0.472 & \textbf{0.149} & \textbf{0.121} & 0.411 \\
V-JEPA-2-L & 0.356 & 0.294 & 0.387 & 0.375 & 0.566 & \textbf{0.484} & 0.119 & 0.000 & 0.402 \\

\bottomrule
\end{tabular}

\caption{\scriptsize Backbone macro-F1 performance on \textbf{CoralCam} and \textbf{FishFollow}.}
\label{tab:backbone_f1_combined}

\vspace{-2.5em}

\end{table*}

\vspace{-1\baselineskip}
\subsection{Temporal vs. Static Cues}

Our results on \ours{}, shown in Table \ref{tab:backbone_f1_combined}, reveal several insights about the role of temporal information in fish behavior classification.

First, models that explicitly capture spatio-temporal dynamics generally outperform image-based backbones. Video models such as VideoMAE and V-JEPA-2 achieve the highest scores on both CoralCam and FishFollow. The advantage is particularly pronounced for behaviors with strong temporal cues. For example, on CoralCam, image-based models fail to detect the \textit{Aggression} class, whereas video models achieve substantial performance (0.332 with VideoMAE-L). This highlights the importance of temporal context for recognizing dynamic or interaction-driven behaviors.

At the same time, image-based models remain surprisingly competitive for behaviors that are primarily defined by static visual cues. On CoralCam, image models achieve performance comparable to the best video models on the \textit{Biting} category (e.g., 0.490 with DINOv3-L versus 0.510 with VideoMAE-L). Similarly, on FishFollow, image backbones achieve the highest performance on the \textit{Habitat} category, suggesting that appearance-based cues alone can be sufficient for certain behavior categories.

Together, these results demonstrate that fish behavior recognition requires both temporal reasoning and strong appearance representations. While spatiotemporal models provide clear advantages for interaction-driven behaviors, image models can achieve competitive performance on appearance-dominated categories at a fraction of the computational cost. This highlights the multi-faceted nature of \ours{}, where different behavioral categories require fundamentally different types of visual reasoning.

\vspace{-1\baselineskip}
\subsection{Frozen Foundation Models vs Fully Fine-Tuned CNNs}

\begin{table*}[t]
\centering
\scriptsize
\setlength{\tabcolsep}{2pt}
\renewcommand{\arraystretch}{1.05}

\begin{tabular}{lccc|cccccc}
\toprule
& \multicolumn{3}{c|}{\textbf{CoralCam}} 
& \multicolumn{6}{c}{\textbf{FishFollow}} \\
\cmidrule(lr){2-4} \cmidrule(lr){5-10}

\textbf{Model} 
& \textbf{Overall} & \textbf{Aggression} & \textbf{Biting}
& \textbf{Overall} & \textbf{Habitat} & \textbf{Movement} & \textbf{Bites} & \textbf{Social} & \textbf{Not Vis.} \\

\midrule

DINOv3-L \\ (Frozen) 
& 0.329 & \textbf{0.007} & 0.490
& \textbf{0.380} & \textbf{0.594} & 0.458 & \textbf{0.123} & \textbf{0.036} & \textbf{0.458} \\

\midrule

ResNet\\ (Frozen) 
& 0.286 & 0.000 & 0.429
& 0.361 & 0.571 & \textbf{0.460} & 0.082 & 0.023 & 0.405 \\

ResNet50\\ (Full) 
& \textbf{0.358} & 0.000 & \textbf{0.537}
& 0.321 & 0.550 & 0.405 & 0.041 & 0.000 & 0.385 \\
\bottomrule
\end{tabular}

\caption{\scriptsize Frozen vs.\ full fine-tuning performance (macro-F1) on \textbf{CoralCam} and \textbf{FishFollow}.}
\label{tab:freeze_vs_full_combined}
\vspace{-2.5em}
\end{table*}

As shown in Table \ref{tab:freeze_vs_full_combined}, on the static and ``simpler" CoralCam dataset, the classical method of fully finetuning a lightweight ResNet50 backbone is surprisingly competitive and even outperforms the modern DINOv3-L. However, this advantage vanishes on the more complex, dynamic, and behaviorally diverse FishFollow dataset (Table \ref{tab:freeze_vs_full_combined}). Here, the fully-tuned ResNet50 appears to overfit, and the robust, frozen features of DINOv3-L are better.
\vspace{-1\baselineskip}
\subsection{Impact of Class Imbalance Strategies}

\begin{table*}[t]
\centering
\scriptsize
\setlength{\tabcolsep}{2.5pt}
\renewcommand{\arraystretch}{1.05}

\begin{tabular}{l c ccc | c ccc}
\toprule
& \multicolumn{4}{c|}{\textbf{Class Imbalance (Focal vs BCE)}} 
& \multicolumn{4}{c}{\textbf{Pooling (Attention vs Mean)}} \\
\cmidrule(lr){2-5} \cmidrule(lr){6-9}

\textbf{Backbone} 
& \textbf{BCE F1} & $\Delta$Precision & $\Delta$Recall & $\Delta$F1
& \textbf{Mean F1} & $\Delta$Precision & $\Delta$Recall & $\Delta$F1 \\
\midrule

\multicolumn{9}{l}{\textbf{CoralCam}} \\
DINOv3-B   & 0.309 & -0.0220 & +0.0358 & -0.0204 & 0.241 & +0.0602 & -0.1263 & +0.0677 \\
DINOv3-L   & 0.325 & -0.0009 & +0.0669 & +0.0042 & 0.238 & +0.0599 & +0.1560 & +0.0916 \\
ResNet50   & 0.285 & +0.0012 & +0.0109 & +0.0006 & 0.250 & +0.0368 & -0.0266 & +0.0360 \\
VideoMAE-B & 0.362 & -0.0986 & -0.0749 & -0.0736 & 0.193 & +0.1835 & -0.1713 & +0.1691 \\
VideoMAE-L & 0.261 & +0.2255 & +0.0423 & +0.1896 & 0.244 & +0.2212 & +0.0532 & +0.2067 \\
V-JEPA-2-L & 0.352 & -0.0099 & +0.0616 & +0.0041 & 0.356 & -0.0792 & -0.1138 & -0.1149 \\
\midrule

\multicolumn{9}{l}{\textbf{FishFollow}} \\
DINOv3-B   & 0.320 & -0.0419 & +0.1860 & +0.0575 & 0.368 & +0.0023 & +0.0382 & +0.0100 \\
DINOv3-L   & 0.373 & -0.0442 & +0.1292 & +0.0065 & 0.365 & +0.0056 & +0.0421 & +0.0149 \\
ResNet50   & 0.361 & -0.0368 & +0.1668 & -0.0002 & 0.350 & +0.0333 & -0.0630 & +0.0110 \\
VideoMAE-B & 0.358 & -0.0563 & +0.1545 & +0.0122 & 0.339 & +0.0061 & +0.1116 & +0.0313 \\
VideoMAE-L & 0.369 & -0.0346 & +0.1606 & +0.0175 & 0.355 & -0.0165 & +0.1308 & +0.0320 \\
V-JEPA-2-L & 0.356 & -0.0930 & +0.1209 & +0.0192 & 0.375 & -0.0325 & +0.0799 & -0.0036 \\
\bottomrule
\end{tabular}

\caption{\textbf{Ablation summary with anchors.} Macro-F1 anchors are \textbf{BCE} (left block) and \textbf{Mean pooling} (right block). $\Delta$ values denote metric differences for \textbf{Focal$-$BCE} and \textbf{Attention$-$Mean} respectively (positive means the first option improves over the anchor).}
\label{tab:ci_and_pooling_anchors}
\vspace{-2em}

\end{table*}

As shown in Table \ref{tab:ci_and_pooling_anchors}, we compare two training strategies for handling severe class imbalance: (1) balanced sampling with standard binary cross-entropy (BCE), and (2) balanced sampling combined with focal loss. Across most configurations, focal loss improves performance by boosting recall at a modest cost to precision. The only case where this does not occur is with VideoMAE on the CoralCam dataset. We suspect focal loss does not help in this case because it caused the model to overfit to very rare aggression behaviors resulting in zero performance on the test set. Similarly, naive random sampling (shown in the appendix) yields near-zero performance on rare behaviors, emphasizing the necessity of targeted sampling strategies for extreme imbalance in \ours{} and other ecological datasets.
\vspace{-1.25\baselineskip}
\subsection{Effect of Pooling Strategies}

As shown in Table \ref{tab:ci_and_pooling_anchors}, attention pooling outperforms mean pooling across nearly all backbones. Attention pooling introduces learnable parameters that allow the model to adaptively weight token representations, enabling the classifier to emphasize more informative spatial or temporal regions.

An exception is V-JEPA-2-L, where mean pooling performs substantially better. We hypothesize that this difference stems from the model's self-supervised pretraining objective. V-JEPA is trained using a predictive embedding objective, which encourages information to be distributed across tokens rather than concentrated in a small subset of salient patches. In this setting, a simple cross-attention pooling mechanism may overemphasize individual tokens and discard useful distributed information, whereas mean pooling aggregates signals across all tokens more robustly.

For consistency across backbones, we report results using the same simple pooling mechanisms. However, because V-JEPA is designed to use a more specialized attentive pooling module, we include experiments with a stronger V-JEPA-specific adaptor in the appendix.
\vspace{-1\baselineskip}
\subsection{Parameter Count and Performance Tradeoff}

\begin{figure*}[t] 
    \centering
    \begin{subfigure}[b]{0.24\textwidth}
        \centering
        \includegraphics[width=\textwidth]{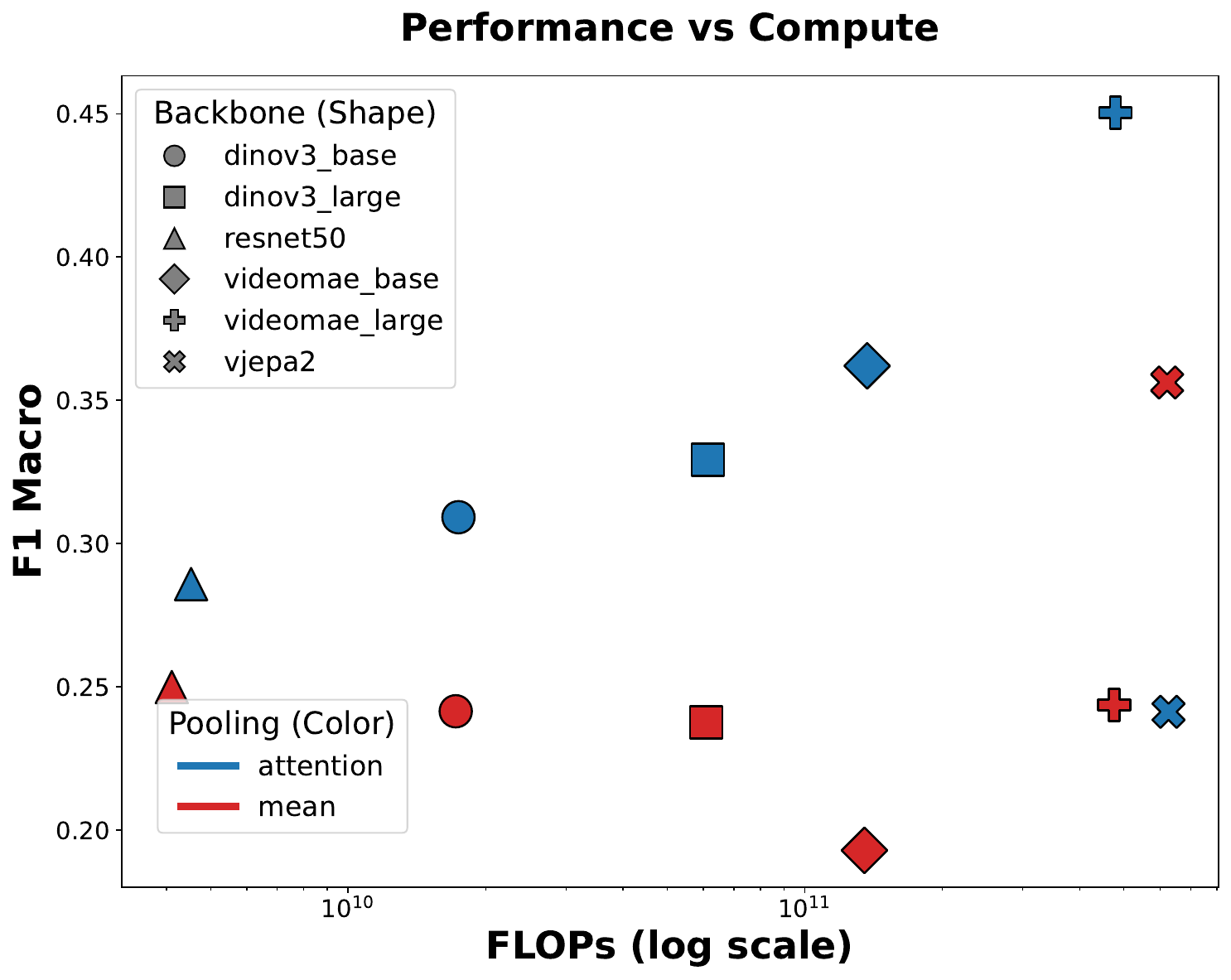}
        \caption{Coral: FLOPs}
    \end{subfigure}
    \hfill
    \begin{subfigure}[b]{0.24\textwidth}
        \centering
        \includegraphics[width=\textwidth]{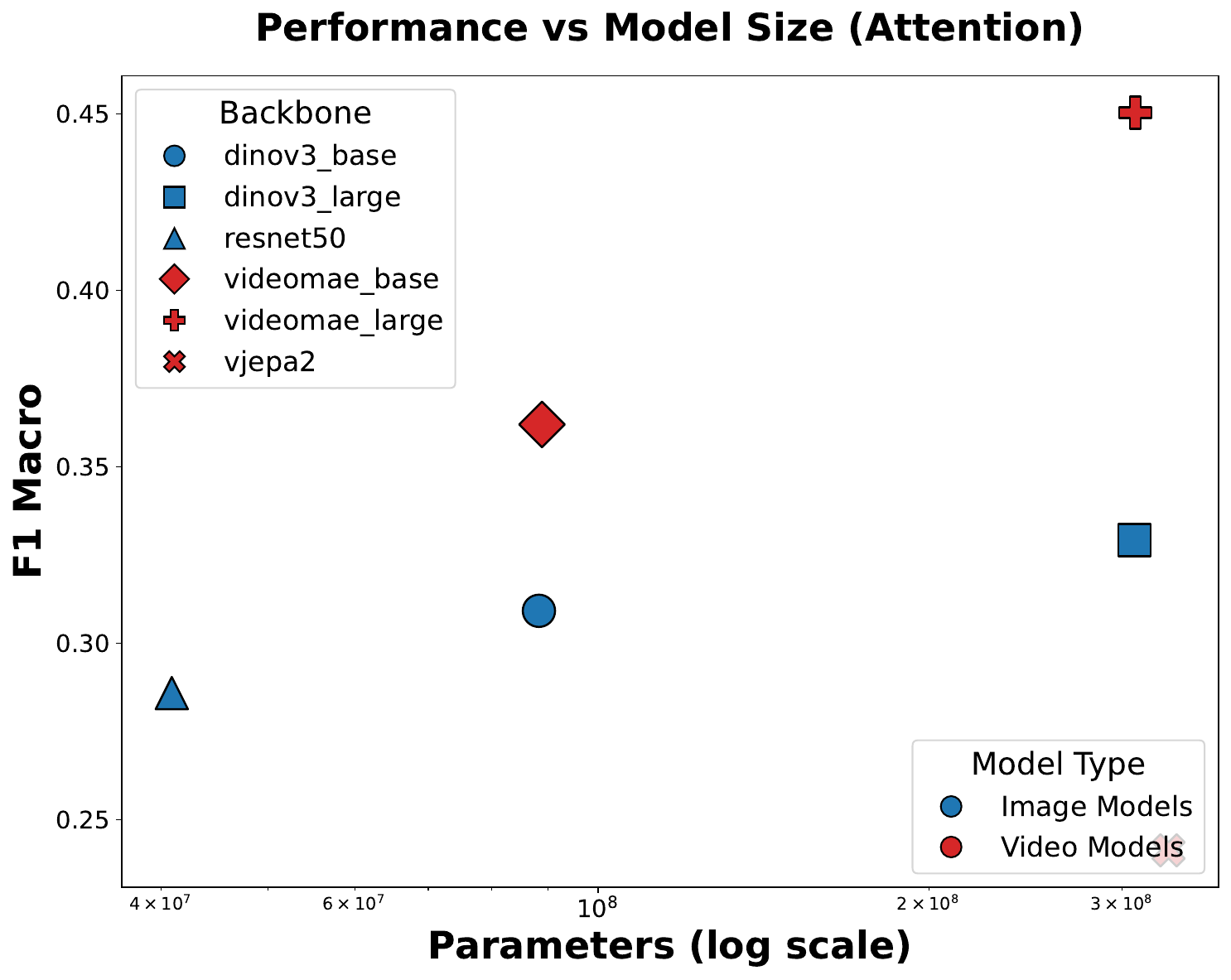}
        \caption{Coral: Params}
    \end{subfigure}
    \hfill
    \begin{subfigure}[b]{0.24\textwidth}
        \centering
        \includegraphics[width=\textwidth]{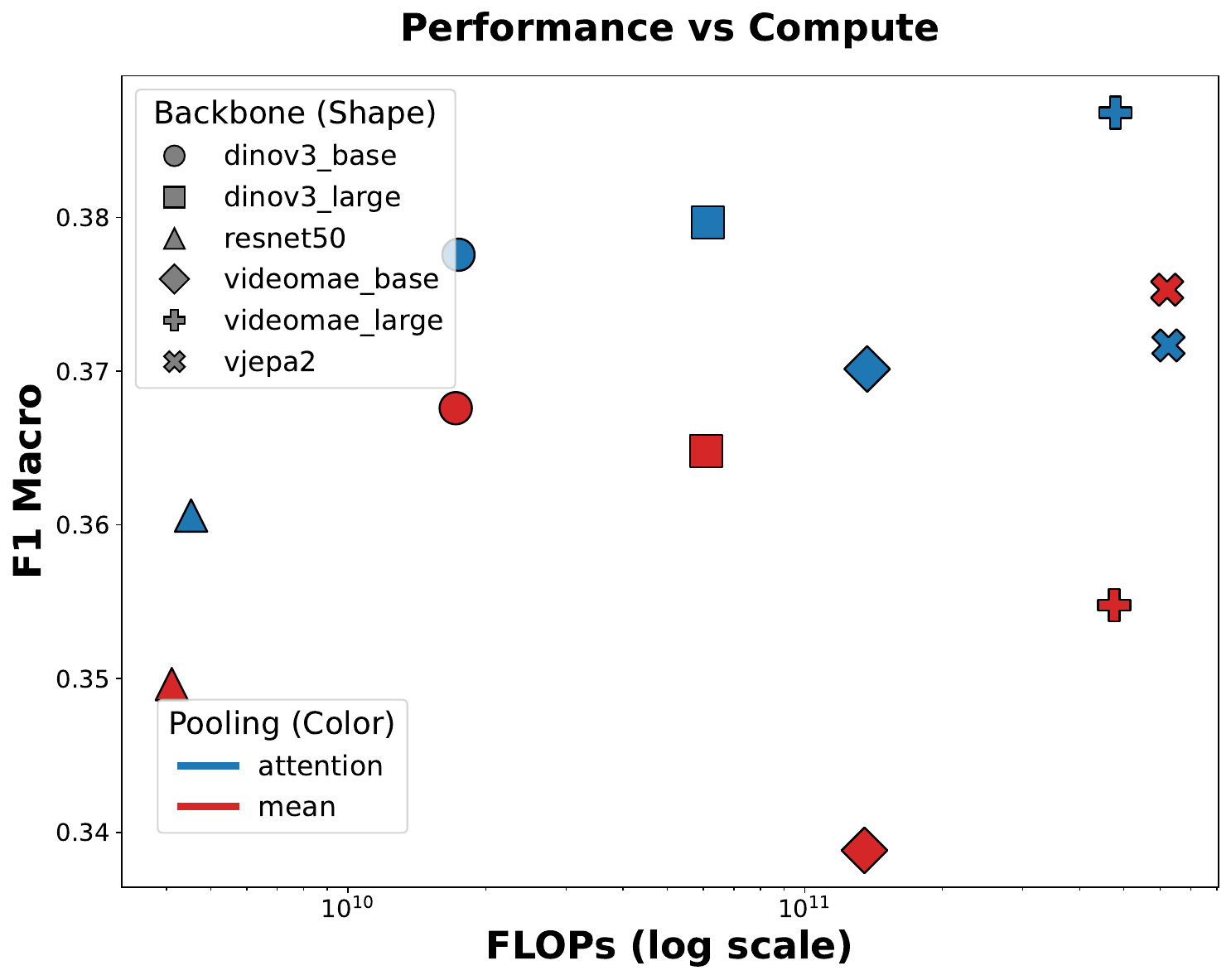}
        \caption{Fish: FLOPs}
    \end{subfigure}
    \hfill
    \begin{subfigure}[b]{0.24\textwidth}
        \centering
        \includegraphics[width=\textwidth]{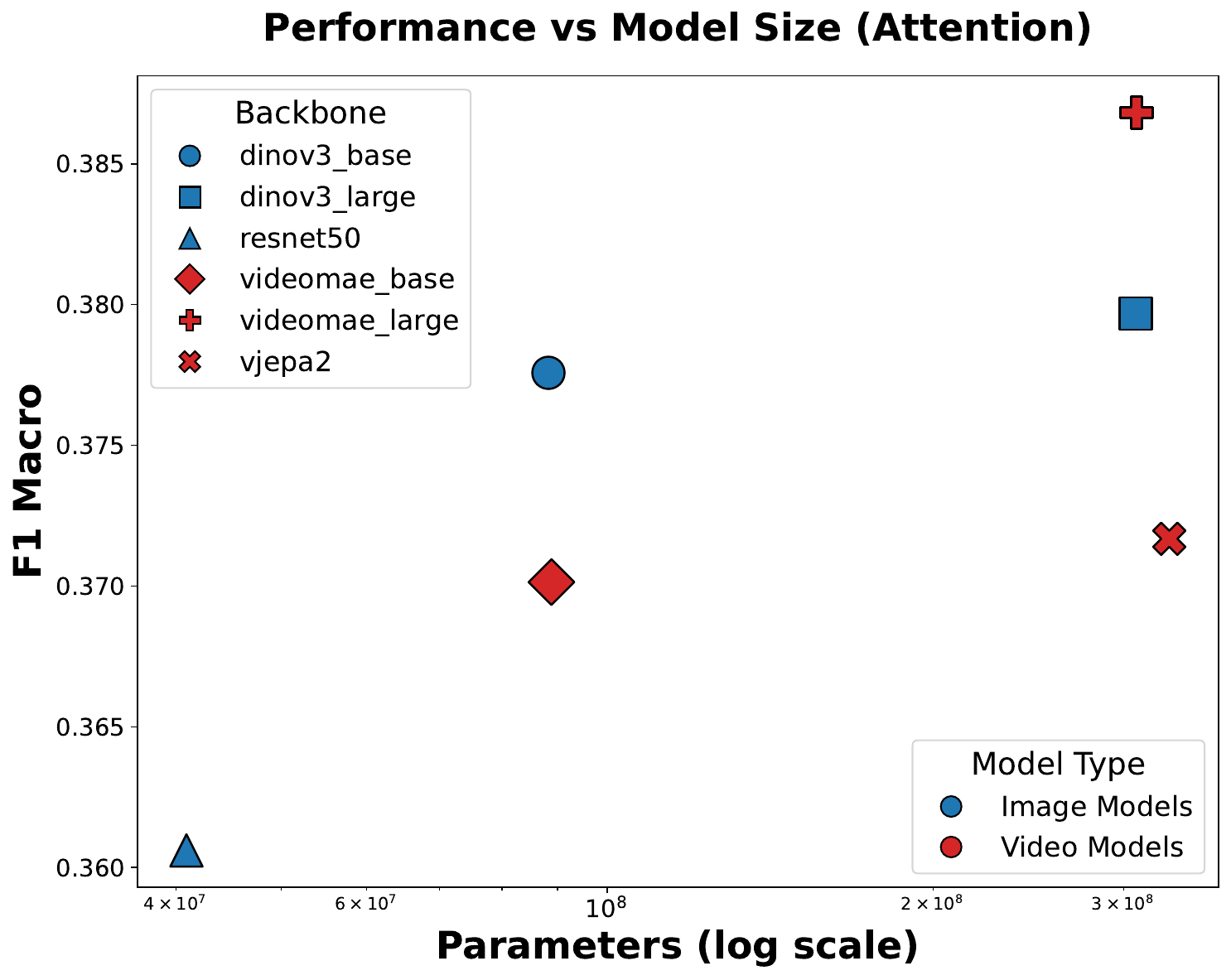}
        \caption{Fish: Params}
    \end{subfigure}
    
    \caption{\scriptsize Compute and performance tradeoffs for CoralCam and FishFollow. (a, c) show macro-F1 vs. FLOPs, while (b, d) show macro-F1 vs. total parameters.}
    \label{fig:performance}
    \vspace{-2em}
\end{figure*}

As shown in Figure~\ref{fig:performance}, attention-based pooling (blue) consistently outperforms mean pooling (red) across nearly all compute scales. In CoralCam   (a), attention pooling exhibits a steady monotonic increase in macro-F1 as FLOPs scale. In contrast, FishFollow   (c) displays higher volatility. Across both datasets, V-JEPA models follow these broader performance gaps between pooling strategies, though they exhibit distinct scaling trajectories compared to the other backbones.

Regarding model size (b, d), video-native backbones (red) generally achieve higher F1 scores than image-based models (blue) for equivalent parameter counts, a trend particularly evident in the challenging CoralCam  environment. Notably, VideoMAE Large achieves peak macro-F1 in both datasets despite its smaller footprint compared to DINOv3 variants. These results suggest that temporal pre-training or video-specialized architectures provide superior parameter efficiency for ecological monitoring tasks compared to general-purpose image backbones.
\vspace{-1\baselineskip}
\subsection{Detection Baselines}

\label{sec:detection-benchmarking}

\begin{wraptable}{r}{0.6\linewidth}
    \centering
    \scriptsize
    \setlength{\tabcolsep}{3pt}
    \renewcommand{\arraystretch}{1.1}

    \vspace{-5em}

    \begin{tabular}{lccc}
    \toprule
    \textbf{Model} & \textbf{AP@0.5} & \textbf{AP@0.75} & \textbf{mAP@[0.5-0.95]} \\
    \midrule
    MobileNet-V3 & 0.205 & 0.051 & 0.082 \\
    ResNet50-V2 & 0.657 & 0.298 & 0.338 \\
    RT-DETR-L & \textbf{0.807} & \textbf{0.525} & \textbf{0.477} \\
    YOLOv11-N & 0.654 & 0.393 & 0.368 \\
    YOLOv11-X & 0.764 & 0.516 & 0.455 \\
    \bottomrule
    \end{tabular}

    \caption{\scriptsize \textbf{Detection performance on the \ours{} detection subset.} 
    Average Precision (AP) reported at IoU thresholds of 0.5, 0.75, and averaged over the range 0.5 to 0.95 with 0.05 increments. MobileNet-V3 and ResNet50-V2 are different feature extractors for Faster R-CNN.}
    \label{tab:detection}
    \vspace{-2em}
\end{wraptable}

To establish a baseline on \ours{}'s object detection dataset, we benchmark several representative detection architectures, ranging from classical CNN-based detectors (Faster R-CNN) to modern transformer and real-time architectures (RT-DETR and YOLOv11). 
As shown in Table~\ref{tab:detection}, the transformer-based RT-DETR-L achieves the highest performance, with YOLOv11-X performing comparably. 
While these models achieve strong results, there remains significant room for improvement. Improving the models and methods on this dataset and the task of in-the-wild fish detection would substantially reduce annotation effort for marine ecological monitoring and downstream behavior labeling.

\vspace{-1.5em}
\section{Conclusion}
\vspace{-1em}
\label{sec:conclusion}

We present \ours, a dataset of in-situ fish behavior. The diverse, expert-annotated behaviors in \ours~have the potential to help shed new light on behavioral dynamics of fish, which are critical to the health of marine ecosystems worldwide \cite{gil2020fast,blackwood2012effect,morais2019pelagic}. 
More generally, the dataset offers a rich range of camera perspectives, background attributes, lighting conditions, subject diversity, and varied behavior, all of which are features that make it particularly challenging to automate analysis of imagery from the field \cite{belcher2023demystifying}. 
We aim for \ours~ as a platform towards interdisciplinary collaboration and methods development in computer vision that will enable automated quantification of animal behavior from underwater imagery and improve overall understanding of fish behavior. Such methods could greatly accelerate analysis of scientific imagery, and add tremendous value to underwater footage collected by recreational divers and citizen scientists, opening the possibility of vastly accelerating data collection from Earth's oceans, rivers, and coral reefs.

\subsection*{Acknowledgments}

This work was partly funded by NSF IIS-2505098 (JJS), NSF IOS-2338569 (AMH), and NSF EF-2222478 (AMH). Research supported by the NVIDIA Academic Grant Program using 4 $\times$ RTX PRO 6000 Blackwell GPUs. We would like to thank Xinyu Yang and the rest of the members of the PIs' labs for helpful discussions and assistance in annotation efforts. We would also like to thank Benjamin Martin and Lars Koopmans for their assistance in the field. 

\bibliographystyle{splncs04}
\bibliography{main}

\clearpage
The sections of our appendix are organized as follows:

\begin{itemize}
    \item Section \ref{supp:dataset-details}: Dataset Details.
    \item Section \ref{supp:experiment-details} Experiment Details.
    \item Section \ref{supp:additional-experiments}: Additional Experimental Results.
    \item Section \ref{supp:datasheet}: Datasheets.
\end{itemize}

\section{Dataset Details}

\label{supp:dataset-details}
In this section, we provide more detailed documentation of the two datasets included in our benchmark: \textbf{CoralCam} and \textbf{FishFollow}. For each dataset, we describe the recording conditions, annotation protocols, and labeling challenges encountered during curation. We also include an ethogram (Table ~\ref{tab:ethogram}) that outlines the behavioral taxonomy used for annotation in CoralCam and FishFollow. These details are intended to support reproducibility, clarify the scope of behavioral coverage, and guide future extensions or applications of the datasets.

\begin{table*}[htbp]
\centering
\renewcommand{\arraystretch}{1.35}

\resizebox{\textwidth}{!}{%
\begin{tabular}{|>{\raggedright\arraybackslash}p{2.2cm}|
                >{\centering\arraybackslash}p{3cm}|  
                >{\centering\arraybackslash}p{3cm}|   
                >{\raggedright\arraybackslash}p{6cm}|}
\hline
\textbf{Dataset} & \textbf{Definition} & \textbf{Subcategory} & \textbf{Subcategory Definition} \\
\hline

\multirow{4}{*}{\textbf{CoralCam}}& 
\multirow{4}{=}{\centering Videos focused on mixed-species schools of reef fish. }
& Biting / Not Biting & Mouth is open and individual is feeding. \\
\cline{3-4}
&& Not Visible & Behavior unable to be annotated. \\
\cline{3-4}
&& C-turn & Individual has initiated a startle response, bending their body in the shape of a C. \\
\cline{3-4}
&& Being Charged & Individual is being targeted by another fish. \\
\hline

\multirow{20}{*}{\textbf{FishFollow}}& \multirow{20}{=}{\centering Videos of individually followed parrotfish across diverse reef habitats.}
& Rubble Habitat & Individual is interacting with a  rubble-dominated habitat.\\
\cline{3-4}
&& Coral Habitat & Individual is interacting with a coral-dominated habitat.\\
\cline{3-4}
&& Sand Habitat & Individual is interacting with sand-dominated habitat.\\
\cline{3-4}
&& Seafloor Bites & Biting directed toward the seafloor. \\
\cline{3-4}
&& Low Bites & Bites occurring near the bottom of the water column. \\
\cline{3-4}
&& Medium Bites & Bites occurring in mid-water. \\
\cline{3-4}
&& High Bites & Bites occurring near the surface. \\
\cline{3-4}
&& Solo Foraging & Foraging alone. \\
\cline{3-4}
&& Foraging & Searching for food. \\
\cline{3-4}
&& Social Foraging & Foraging in the presence of conspecifics. \\
\cline{3-4}
&& Not Visible & Individual not visible in the frame. \\
\cline{3-4}
&& Cleaner Wrasse & Interaction with a cleaner wrasse. \\
\cline{3-4}
&& Change in Focal Fish & The focal fish has changed. \\
\cline{3-4}
&& Traversing & Swimming through habitat without foraging. \\
\cline{3-4}
&& Departure & Leaving the frame or habitat patch. \\
\cline{3-4}
&& Idle & Remaining stationary without feeding. \\
\cline{3-4}
&& Aggressive on Focal & Another fish directs aggression at the focal fish. \\
\cline{3-4}
&& Aggressive by Focal & Focal fish directs aggression at another. \\
\cline{3-4}
&& Sand Rubbing & Rubbing body on the sand. \\
\cline{3-4}
&& Other & Other unclassified behaviors. \\
\hline

\end{tabular}
}
\caption{Ethogram for CoralCam and FishFollow datasets.}
\label{tab:ethogram}
\end{table*}

\subsection{CoralCam}
\paragraph{Collection and Organization.} The CoralCam dataset is a curated collection of 12 videos of ecological field data.  These videos capture large, mixed-species schools composed primarily of \textit{Azurina multilineata} (brown chromis), \textit{Stegastes partitus} (bicolor damselfish), and juvenile \textit{Thalassoma bifasciatum} (bluehead wrasse). All videos were recorded in December 2023 at Playa Largu, Curaçao. The videos selected for this dataset were based on criteria intended to optimize tracking performance, including low background complexity, minimal reef structures, and relatively sparse fish densities to reduce conspecific occlusion.

Following tracking, 213 individual tracks were selected for behavior annotation. These tracks were chosen based on their length-averaging approximately 1,215 frames—and were manually verified by an expert to ensure that the same individual was consistently tracked throughout each sequence.

\paragraph{Annotation Process and Quality Control.}
 Bounding box annotations used to train object detection and tracking models were produced by a team of five trained student annotators. These annotators were supervised and trained by a single expert, who also conducted the field recordings. Bounding boxes were labeled for all three target species and reviewed by the expert to ensure consistency and accuracy. In total, 21,116 bounding boxes were annotated and verified.

Once tracking was completed, the 213 verified tracks were selected for frame-by-frame behavior annotation. All behavior annotations were performed by a single expert using a predefined behavior schema (Table ~\ref{tab:ethogram}). Note that some of these behaviors are excluded from our evaluation because they occur so infrequently (less than 30 examples in the training set). On CoralCam, the \textit{C-Turn} class was removed from evaluation. On FishFollow, the classes that were removed are \textit{Other Behavior}, \textit{Departure}, \textit{Change in Focal Fish}, \textit{Aggressive on Focal}, \textit{Aggressive by Focal}, \textit{Sand Rubbing}, and \textit{High Bites}. Although not used in our main analyses, these rare categories may still be valuable for future retrieval or retrieval-augmented tasks.

\begin{figure*}[t]
    \centering
    \includegraphics[width=1.0\linewidth]{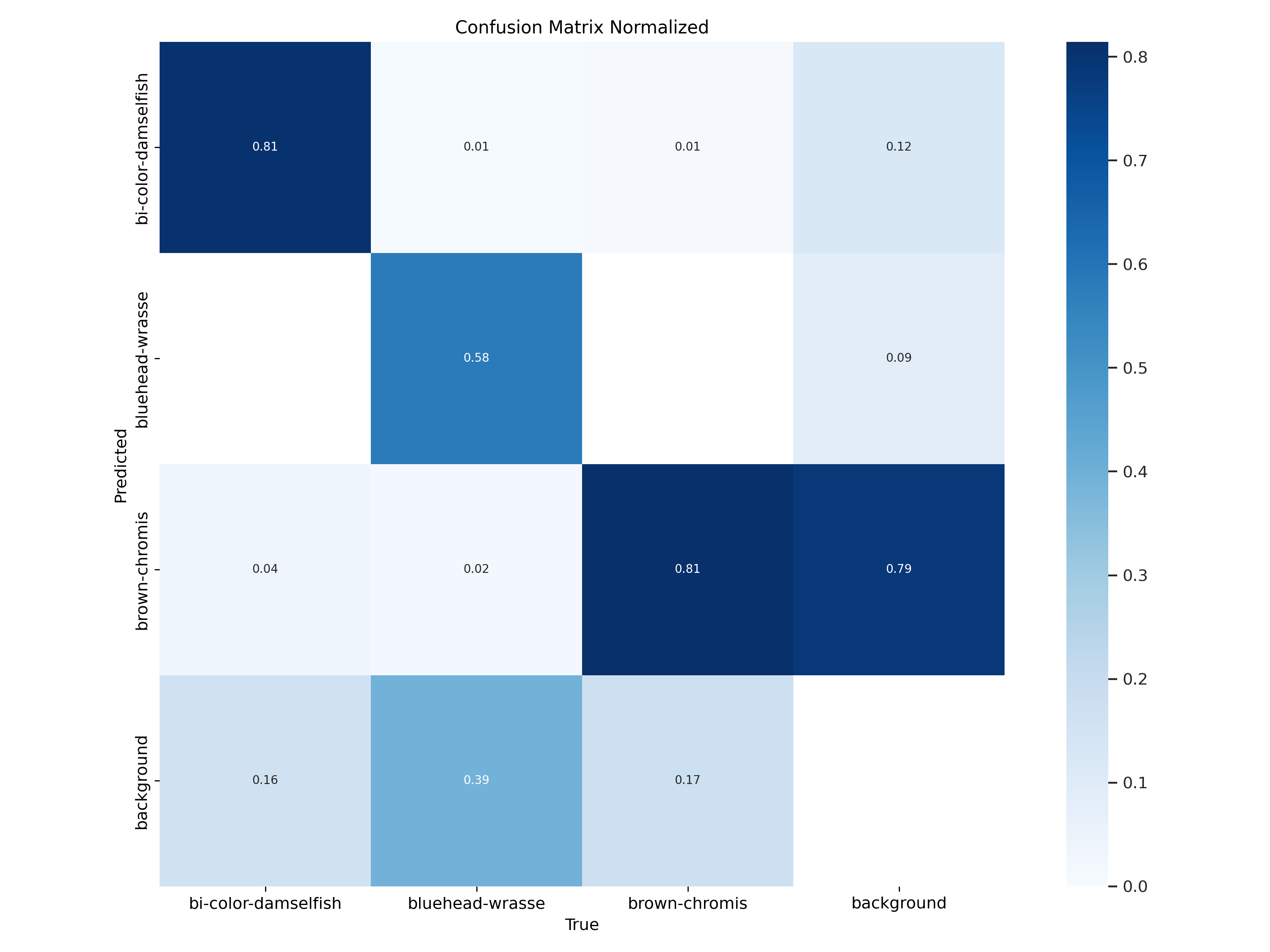}

    \caption{\scriptsize \textbf{Coral Cam Object Detector, Normalized Confusion matrix: } Column-normalized over the true (ground-truth) class, so each column sums to one. The detector recovers brown chromis and bi-color damselfish reliably (recall 0.81 each) but misses a substantial fraction of bluehead wrasse (0.39 assigned to background). The largest off-diagonal entry, 0.79 of background predicted as brown chromis, reflects false-positive brown-chromis detections; because CoralCam footage is dominated by dense brown-chromis (A. multilineata) schools of which only a subset was annotated, these are most plausibly correct detections of distant, unannotated individuals rather than true detector errors, and thus partly an artifact of annotation incompleteness. Inter-species confusion is otherwise minimal (all species–species off-diagonals $\leq$ 0.04).
 }
    \label{fig:confusion}

\end{figure*}

Bounding box annotations were used to train and test an object detector with a YOLOv8 backbone, which achieved a mAP@50 score of 74.5 of the test set. Performance of the model is higher for the primary species of interest (brown chromis). Common mistakes include background detections, which can often be attributed to animals far away from the camera. A normalized confusion matrix is included to illustrate model performance by class (Fig.~\ref{fig:confusion}).  This object detector was subsequently deployed on videos from each of the 51 sites, generating detections for each frame in the dataset. Tracks for each detection were generated using the BoTsort algorithm \cite{aharon2022botsortrobustassociationsmultipedestrian}. Each frame typically contains tens to hundreds of individuals, making frame-level behavioral annotations impossible for full scenes, as individuals may exhibit distinct behaviors. To address this, we used multi-object tracking to extract numerous tracks per video, each isolating a single focal individual. These tracks were converted into clips, enabling frame-level behavioral annotations. Model weights were generated using the work of a single expert annotator, however, a small subset of these videos were also annotated by an expert trained annotator for comparison. The F1 score calculated among the two annotators was 0.74.

\paragraph{Tackling Annotation Challenges.}
Annotators encountered several challenges during labeling. The most significant issue was the cryptic appearance of focal individuals against the complex reef background, which made detection and tracking difficult. This was mitigated through manual expert verification of both bounding boxes and track identities. Additionally, fish orientation occasionally obscured behavioral cues, making behavior classification impossible in some frames. To address this, a “not visible” behavior class was introduced to capture moments when individuals were facing away from the camera, out of frame, or otherwise occluded. 

\paragraph{Summary of data}. The data made available within CoralCam can be summarized as follows:
\begin{itemize}
    \item 12 videos
    \item 10–20 tracked individuals per video
    \item \verb|.txt| files containing annotations for each track
    \item \verb|.pkl| files containing bounding box information for each track
    \item YOLO model weights trained for fish detection and tracking
    \item Training images in COCO format for the custom-trained model
\end{itemize}

\paragraph{Sample data}
We include a sample video visualizing sample tracks in the context of the full video. The punch outs pictured in the sample video are characteristic of the individual tracks which a researcher annotates.  Together, this sample data illustrates the complexity of the fish schools and the range of behaviors characteristic of this system

\subsection{FishFollow}

\paragraph{Collection and Annotation.}
The FishFollow dataset comprises 8 hours of video footage recorded in Curaçao during June and July 2023. Data were collected by a team of two snorkelers, each tasked with identifying and following a single parrotfish individual for approximately 20 minutes. Because individuals moved freely throughout their environment, the footage captures a broad range of habitats—including open reef and rubble—and a diverse set of behavioral and social contexts, such as foraging and interactions with conspecifics.

\paragraph{Annotation Process and Quality Control.}
 Annotation was conducted by a group of undergraduate students trained in fish behavior identification. Each student received a reference video, created by an expert, containing examples of each target behavior. Each video in the data set was independently annotated by three different students. Annotations were time-based and conducted using BORIS, an open-source software widely used for behavioral studies. These time-based labels were subsequently converted to frame-by-frame annotations. Upon completion, one expert reviewed and confirmed the accuracy of all behavior classifications.

 Average precision of annotations is 0.84 and average recall is 0.75 among all annotator pairs. When the tolerance is increased to 0.5 seconds (\~30 frames), the average F1 score increases to 0.86 with a median of 0.916. Furthermore, increased tolerance results in an average precision of 0.925 and recall of 0.84. A higher tolerance is acceptable in the case of FishFollow, since many behaviors have long transition times and the edges of behaviors are not necessarily clear. A total of 12 annotators were included in this dataset.

\paragraph{Tackling Annotation Challenges.}
 Parrotfish are highly camouflaged and often blend into their surroundings, which posed challenges in consistently identifying the focal individual and accurately labeling behavior. This challenge was addressed by implementing redundant labeling (i.e., multiple annotators per video) and conducting a thorough expert review. Additionally, students were provided with reference clips illustrating each behavior class to promote consistent and reliable labeling. In addition to \ref{tab:ethogram}, we have include a video ethogram which was used in training annotators, and includes a wide variety of expected behavior. 

\section{Experiment Details}

\label{supp:experiment-details}
\subsection{Train/Test Split Details}

To evaluate model performance fairly, we partition each dataset into non-overlapping training and test sets. Approximately 70\% of the data is used for training and 30\% for testing, with each video assigned to only one split to prevent information leakage across sets.

For each dataset, a validation set is further constructed by randomly selecting ~10\% of the training set using the same split criteria. This ensures the validation data is representative while avoiding overlap with the test set.

Tables~\ref{tab:coralcam_split} and~\ref{tab:fish_split_behavior} summarize the number of files, total frames, and per-frame behavior frequencies for the CoralCam and FishFollow datasets, respectively.

\begin{table*}[ht]
\centering
\begin{tabular}{lcccccccc}
\toprule
 \textbf{Split}&\textbf{Files}  & \textbf{Frames} &  \textbf{Mouth Not Visible} &\textbf{Bite}&\textbf{Being Charged}&\textbf{C Turn}\\
\midrule
 \textbf{Train}& 150 &184178&  0.11 &0.0084  &0.00092 &0.00015 \\

 \textbf{Test}&  63&74820& 0.1 &0.0088  &0.00056 &0.00011  \\

 \textbf{Overall}& 213 &258998& 0.1 &0.0085  &0.00081 &0.00014   \\
\bottomrule
\end{tabular}
\caption{Coral Cam Train/Test Split Size and Behavior Frequency.}

\label{tab:coralcam_split}
\end{table*}

\begin{table*}[ht]
\centering
\scriptsize

\resizebox{0.98\linewidth}{!}{
\begin{tabular}{lcccccccc}
\toprule
\textbf{Split} & \textbf{Files} & \textbf{Frames} &
\textbf{Other behavior} & \textbf{Medium Bites} & \textbf{High Bites} &
\textbf{Traversing} & \textbf{Departure} & \textbf{Cleaner Wrasse} \\
\midrule
\textbf{Train}   & 57 & 1265160 & 2.4e-06 & 0.00021 & 0.00023 & 0.28 & 6.2e-05 & 0.019 \\
\textbf{Test}    & 24 & 534714  & 1.1e-05 & 0.00037 & 3.2e-05 & 0.22 & 3.2e-05 & 0.031 \\
\textbf{Overall} & 81 & 1799894 & 5e-06   & 0.00026 & 0.00017 & 0.27 & 5.3e-05 & 0.023 \\
\bottomrule
\end{tabular}
}

\resizebox{0.98\linewidth}{!}{
\begin{tabular}{lcccccccc}
\toprule
\textbf{Split} &
\textbf{Low Bites} & \textbf{Change in Focal Fish} &
\textbf{Solo Foraging} & \textbf{Seafloor Bites} &
\textbf{Aggressive on Focal} & \textbf{Social Foraging} &
\textbf{Idle} \\
\midrule
\textbf{Train}   & 0.0017 & 3.2e-06 & 0.68 & 0.0039 & 5.1e-05 & 0.14 & 0.0057 \\
\textbf{Test}    & 0.0019 & 3.7e-06 & 0.62 & 0.0039 & 3.2e-05 & 0.13 & 0.0078 \\
\textbf{Overall} & 0.0017 & 3.3e-06 & 0.66 & 0.0039 & 4.5e-05 & 0.14 & 0.0063 \\
\bottomrule
\end{tabular}
}

\resizebox{0.98\linewidth}{!}{
\begin{tabular}{lccccccc}
\toprule
\textbf{Split} &
\textbf{Foraging} &
\textbf{Coral Habitat} & \textbf{Rubble Habitat} &
\textbf{Aggressive by Focal} & \textbf{Sand Habitat} &
\textbf{Sand Rubbing} & \textbf{Not Visible} \\
\midrule
\textbf{Train}   & 0.49 & 0.33 & 0.47 & 1.6e-05 & 0.031 & 1.5e-05 & 0.19 \\
\textbf{Test}    & 0.48 & 0.17 & 0.54 & 7.5e-06 & 0.05  & 1.3e-05 & 0.26 \\
\textbf{Overall} & 0.49 & 0.29 & 0.49 & 1.3e-05 & 0.037 & 1.4e-05 & 0.21 \\
\bottomrule
\end{tabular}
}

\caption{Fish Follow Train/Test Split Size and Behavior Frequency.}

\label{tab:fish_split_behavior}
\end{table*}

\subsection{Training Method Details}

\paragraph{Compute resources and feature extraction.}  
We adopt a two-stage training pipeline. First, we freeze the backbone and pre-extract features from each sampled frame or 16-frame video clip. This step is parallelized across multiple processes, with each video or track processed independently. Feature extraction for a single frozen backbone roughly takes $3$ GPU hours for CoralCam and $20$ GPU hours for FishFollow, although specific time estimates depend on backbone sizes and hardware conditions. 

After feature extraction, we train lightweight classification head on top of the frozen features. This training phase required approximately $1 \sim 10$ GPU hours per experiment, and final inference on the full testing set required $1$ GPU hour per pass. All experiments were conducted on NVIDIA A6000 or H100 GPUs using PyTorch Lightning.

\paragraph{Training protocol and convergence.}  
We trained all models using the recommended hyperparameters from the original DINOv3, VideoMAE, V-JEPA-2, and ResNet50 implementations. Each model was trained for 40 epochs which was chosen as it was a sufficiently large number of epochs to ensure convergence across all configurations. Rather than early stopping, we selected the checkpoint with the highest validation mAP for final evaluation on the test set.

\paragraph{Input preprocessing.}  
After sampling, each input is resized and center-cropped to match the input requirements of the foundation models. Image-based models process inputs of shape $(3, 224, 224)$, while video-based models receive 16-frame clips of shape $(16, 3, 224, 224)$. These transformations are applied consistently across the training, validation, and test phases.

\paragraph{Label encoding.}  
CoralCam and FishFollow are multi-label datasets, with each frame or clip possibly containing multiple simultaneous behaviors. We encode each label as a multi-hot vector, indicating the presence or absence of each behavior class. At test time, we apply a tolerance region (e.g., $\pm7$ frames) to accommodate timing uncertainty in short-duration behavioral annotations.

\subsection{Object Detection Details}

All models were trained until convergence using the default settings from their public implementations, with only minimal hyperparameter tuning performed on the validation split. This ensured a fair and reproducible comparison across architectures without favoring any single model through extensive per-model optimization. All experiments were conducted on NVIDIA A6000 or H100 GPUs and took up to 2 GPU Hours.

Final performance was reported on the held-out test split. Following COCO evaluation protocols, we compute Average Precision (AP) at IoU thresholds of 0.50 and 0.75, as well as the mean AP across IoU thresholds from 0.50 to 0.95 in steps of 0.05 (\texttt{mAP@[.5:.95]}).

\section{Additional Experimental Results}
\label{supp:additional-experiments}
In this section, we provide the comprehensive set of experimental results on \ours{}. These include experiments from all combinations of backbone, pooling, and class imbalance strategies for CoralCam in Table \ref{tab:coralcam_perclass_results} and FishFollow in Table \ref{tab:fishfollow_results}. We also record the results using mAP as a metric for both splits in Table \ref{tab:coralcam_map_perclass_results} for CoralCam and Table \ref{tab:fishfollow_map_results} for FishFollow. We report per class scores for CoralCam, but retain the grouped scores on FishFollow for legibility. All per class experimental results are also available in the public codebase.

We ablate the stronger V-JEPA-2 adaptor that is used in the original V-JEPA-2 paper's code. These results are denoted as `V-Attn'. We see that while the results generally improve over simple attention pooling, they still fall short of mean pooling. We also ablate uniform sampling and see that it actually achieves comparable overall performance. We see in our results that uniform sampling often achieves better performance on behaviors that are more frequent at the cost of poorer performance on rarer behaviors. 

\begin{table*}[t]
\centering
\scriptsize
\setlength{\tabcolsep}{5pt}
\renewcommand{\arraystretch}{1.05}

\begin{tabular}{lcccc}
\toprule
& \multicolumn{4}{c}{\textbf{CoralCam Performance (macro/per-class F1)}} \\
\cmidrule(lr){2-5}

\textbf{Model (Pooling/CI Strategy)}
& \textbf{Overall}
& \textbf{Mouth Not Visible}
& \textbf{Feeding}
& \textbf{Being Charged} \\

\midrule

DINOv3-B (Attention/Focal)
& 0.289 & 0.554 & 0.313 & 0.000 \\
DINOv3-B (Attention/Uniform)
& 0.309 & \textbf{0.588} & 0.339 & 0.000 \\
DINOv3-B (Mean/Focal)
& 0.241 & 0.491 & 0.228 & 0.005 \\
DINOv3-B (Mean/Uniform)
& 0.026 & 0.000 & 0.078 & 0.000 \\

\midrule

DINOv3-L (Attention/Focal)
& 0.329 & 0.534 & 0.447 & 0.007 \\
DINOv3-L (Attention/Uniform)
& 0.325 & 0.586 & 0.389 & 0.000 \\
DINOv3-L (Mean/Focal)
& 0.230 & 0.459 & 0.230 & 0.001 \\
DINOv3-L (Mean/Uniform)
& 0.238 & 0.444 & 0.268 & 0.001 \\

\midrule

ResNet50 (Attention/Focal)
& 0.286 & 0.504 & 0.354 & 0.000 \\
ResNet50 (Attention/Uniform)
& 0.285 & 0.546 & 0.309 & 0.000 \\
ResNet50 (Mean/Focal)
& 0.225 & 0.441 & 0.233 & 0.000 \\
ResNet50 (Mean/Uniform)
& 0.250 & 0.479 & 0.271 & 0.000 \\

\midrule

VideoMAE (Attention/Focal)
& 0.288 & 0.506 & 0.285 & 0.074 \\
VideoMAE (Attention/Uniform)
& 0.362 & 0.518 & 0.257 & 0.311 \\
VideoMAE (Mean/Focal)
& 0.193 & 0.335 & 0.175 & 0.069 \\
VideoMAE (Mean/Uniform)
& 0.141 & 0.250 & 0.089 & 0.085 \\

\midrule

VideoMAE-L (Attention/Focal)
& \textbf{0.450} & 0.532 & \textbf{0.488} & \textbf{0.332} \\
VideoMAE-L (Attention/Uniform)
& 0.261 & 0.436 & 0.260 & 0.086 \\
VideoMAE-L (Mean/Focal)
& 0.230 & 0.412 & 0.214 & 0.065 \\
VideoMAE-L (Mean/Uniform)
& 0.244 & 0.345 & 0.271 & 0.116 \\

\midrule

V-JEPA2 (Attention/Focal)
& 0.097 & 0.259 & 0.031 & 0.003 \\
V-JEPA2 (Attention/Uniform)
& 0.241 & 0.461 & 0.226 & 0.037 \\
V-JEPA2 (Mean/Focal)
& 0.356 & 0.513 & 0.262 & 0.294 \\
V-JEPA2 (Mean/Uniform)
& 0.352 & 0.572 & 0.298 & 0.186 \\
V-JEPA2 (V-Attn/Focal)
& 0.271 & 0.494 & 0.265 & 0.056 \\
V-JEPA2 (V-Attn/Uniform)
& 0.343 & 0.570 & 0.302 & 0.156 \\

\bottomrule
\end{tabular}

\caption{\scriptsize Performance comparison of backbones, pooling, and CI strategies on the
\textbf{CoralCam} dataset. Overall reports macro-F1 across the three classes; the remaining
columns report the corresponding per-class F1 scores.}
\label{tab:coralcam_perclass_results}
\vspace{-2em}
\end{table*}

\begin{table*}[ht]
\centering
\scriptsize
\setlength{\tabcolsep}{3.5pt}
\renewcommand{\arraystretch}{1.05}

\begin{tabular}{lcccccc}
\toprule
& \multicolumn{6}{c}{\textbf{FishFollow Performance (macro-F1)}} \\
\cmidrule(lr){2-7}

\textbf{Model (Pooling/CI Strategy)} 
& \textbf{Overall} & \textbf{Habitat} & \textbf{Movement} & \textbf{Bites} & \textbf{Social} & \textbf{Not Vis.} \\

\midrule

DINOv3-B (Attention/Focal) & 0.378 & 0.582 & 0.465 & 0.108 & 0.055 & 0.457 \\
DINOv3-B (Attention/Uniform)  & 0.320 & 0.577 & 0.444 & 0.002 & 0.013 & 0.189 \\
DINOv3-B (Mean/Focal) & 0.368 & 0.560 & 0.464 & 0.091 & 0.135 & 0.369 \\
DINOv3-B (Mean/Uniform)  & 0.317 & 0.568 & 0.441 & 0.001 & 0.021 & 0.193 \\

\midrule

DINOv3-L (Attention/Focal) & 0.380 & 0.594 & 0.458 & 0.123 & 0.036 & 0.458 \\
DINOv3-L (Attention/Uniform)  & 0.373 & \textbf{0.605} & 0.480 & 0.039 & 0.071 & 0.449 \\
DINOv3-L (Mean/Focal) & 0.365 & 0.566 & 0.458 & 0.111 & 0.035 & 0.388 \\
DINOv3-L (Mean/Uniform)  & 0.317 & 0.575 & 0.432 & 0.005 & 0.039 & 0.187 \\

\midrule

ResNet50 (Attention/Focal) & 0.360 & 0.548 & 0.454 & 0.103 & 0.028 & 0.433 \\
ResNet50 (Attention/Uniform)  & 0.361 & 0.571 & 0.460 & 0.082 & 0.023 & 0.405 \\
ResNet50 (Mean/Focal) & 0.350 & 0.549 & 0.449 & 0.102 & 0.032 & 0.315 \\
ResNet50 (Mean/Uniform)  & 0.323 & 0.546 & 0.425 & 0.000 & \textbf{0.249} & 0.188 \\

\midrule

VideoMAE (Attention/Focal) & 0.370 & 0.552 & 0.454 & 0.105 & 0.085 & 0.486 \\
VideoMAE (Attention/Uniform)  & 0.358 & 0.582 & 0.455 & 0.072 & 0.133 & 0.285 \\
VideoMAE (Mean/Focal) & 0.339 & 0.522 & 0.445 & 0.097 & 0.009 & 0.314 \\
VideoMAE (Mean/Uniform)  & 0.279 & 0.466 & 0.381 & 0.000 & 0.000 & 0.321 \\

\midrule

VideoMAE-L (Attention/Focal)  & \textbf{0.387} & 0.563 & 0.472 & \textbf{0.149} & 0.121 & 0.411 \\
VideoMAE-L (Attention/Uniform)   & 0.369 & 0.574 & 0.465 & 0.131 & 0.024 & 0.334 \\
VideoMAE-L (Mean/Focal)  & 0.355 & 0.541 & 0.467 & 0.111 & 0.038 & 0.285 \\
VideoMAE-L (Mean/Uniform)   & 0.291 & 0.530 & 0.407 & 0.008 & 0.000 & 0.134 \\

\midrule

V-JEPA2 (Attention/Focal)  & 0.372 & 0.552 & 0.471 & 0.108 & 0.003 & \textbf{0.495} \\
V-JEPA2 (Attention/Uniform)   & 0.355 & 0.588 & 0.453 & 0.076 & 0.026 & 0.325 \\
V-JEPA2 (Mean/Focal)  & 0.375 & 0.566 & 0.484 & 0.119 & 0.000 & 0.402 \\
V-JEPA2 (Mean/Uniform)   & 0.356 & 0.560 & \textbf{0.519} & 0.046 & 0.003 & 0.214 \\
V-JEPA2 (V-Attn/Focal)  & 0.350 & 0.547 & 0.445 & 0.080 & 0.000 & 0.447 \\
V-JEPA2 (V-Attn/Uniform)   & 0.349 & 0.551 & 0.460 & 0.056 & 0.000 & 0.415 \\

\bottomrule
\end{tabular}

\caption{\scriptsize Performance comparison of backbones, pooling, and CI strategies on the \textbf{FishFollow} dataset.}
\label{tab:fishfollow_results}
\vspace{-2em}
\end{table*}

\begin{table*}[t]
\centering
\scriptsize
\setlength{\tabcolsep}{5pt}
\renewcommand{\arraystretch}{1.05}

\begin{tabular}{lcccc}
\toprule
& \multicolumn{4}{c}{\textbf{CoralCam Performance (mAP/per-class AP)}} \\
\cmidrule(lr){2-5}

\textbf{Model (Pooling/CI Strategy)}
& \textbf{Overall}
& \textbf{Mouth Not Visible}
& \textbf{Feeding}
& \textbf{Being Charged} \\

\midrule

DINOv3-B (Attention/Focal)
& 0.295 & 0.464 & 0.421 & 0.001 \\
DINOv3-B (Attention/Uniform)
& 0.276 & 0.462 & 0.364 & 0.002 \\
DINOv3-B (Mean/Focal)
& 0.165 & 0.346 & 0.148 & 0.002 \\
DINOv3-B (Mean/Uniform)
& 0.078 & 0.200 & 0.034 & 0.000 \\

\midrule

DINOv3-L (Attention/Focal)
& \textbf{0.322} & 0.482 & \textbf{0.481} & 0.001 \\
DINOv3-L (Attention/Uniform)
& 0.315 & \textbf{0.505} & 0.440 & 0.001 \\
DINOv3-L (Mean/Focal)
& 0.151 & 0.320 & 0.132 & 0.001 \\
DINOv3-L (Mean/Uniform)
& 0.108 & 0.254 & 0.070 & 0.001 \\

\midrule

ResNet50 (Attention/Focal)
& 0.159 & 0.376 & 0.099 & 0.001 \\
ResNet50 (Attention/Uniform)
& 0.161 & 0.393 & 0.090 & 0.001 \\
ResNet50 (Mean/Focal)
& 0.121 & 0.295 & 0.066 & 0.002 \\
ResNet50 (Mean/Uniform)
& 0.107 & 0.269 & 0.050 & 0.002 \\

\midrule

VideoMAE (Attention/Focal)
& 0.119 & 0.325 & 0.027 & 0.004 \\
VideoMAE (Attention/Uniform)
& 0.130 & 0.305 & 0.022 & 0.064 \\
VideoMAE (Mean/Focal)
& 0.112 & 0.191 & 0.019 & 0.126 \\
VideoMAE (Mean/Uniform)
& 0.104 & 0.167 & 0.015 & 0.131 \\

\midrule

VideoMAE-L (Attention/Focal)
& 0.201 & 0.320 & 0.079 & 0.205 \\
VideoMAE-L (Attention/Uniform)
& 0.083 & 0.213 & 0.020 & 0.017 \\
VideoMAE-L (Mean/Focal)
& 0.143 & 0.239 & 0.032 & 0.159 \\
VideoMAE-L (Mean/Uniform)
& 0.127 & 0.194 & 0.027 & 0.160 \\

\midrule

V-JEPA2 (Attention/Focal)
& 0.042 & 0.113 & 0.011 & 0.001 \\
V-JEPA2 (Attention/Uniform)
& 0.100 & 0.272 & 0.022 & 0.007 \\
V-JEPA2 (Mean/Focal)
& 0.252 & 0.454 & 0.041 & \textbf{0.262} \\
V-JEPA2 (Mean/Uniform)
& 0.235 & 0.406 & 0.042 & 0.256 \\
V-JEPA2 (V-Attn/Focal)
& 0.118 & 0.319 & 0.028 & 0.007 \\
V-JEPA2 (V-Attn/Uniform)
& 0.143 & 0.353 & 0.031 & 0.045 \\

\bottomrule
\end{tabular}

\caption{\scriptsize Mean Average Precision (mAP) comparison of backbones and strategies on the
\textbf{CoralCam} dataset. Overall reports mAP across the three classes; the remaining columns
report the corresponding per-class average precision (AP).}
\label{tab:coralcam_map_perclass_results}
\vspace{-2em}
\end{table*}

\begin{table*}[t]
\centering
\scriptsize
\setlength{\tabcolsep}{2pt}
\renewcommand{\arraystretch}{1.05}

\begin{tabular}{lcccccc}
\toprule
& \multicolumn{6}{c}{\textbf{FishFollow Performance (mAP)}} \\
\cmidrule(lr){2-7}

\textbf{Model (Pooling/CI Strategy)} 
& \textbf{Overall} & \textbf{Habitat} & \textbf{Movement} & \textbf{Bites} & \textbf{Social} & \textbf{Not Vis.} \\

\midrule

DINOv3-B (Attention/Focal) & 0.305 & 0.524 & 0.386 & 0.004 & 0.033 & 0.419 \\
DINOv3-B (Attention/Uniform)  & 0.293 & 0.500 & 0.379 & 0.004 & 0.032 & 0.378 \\
DINOv3-B (Mean/Focal) & 0.297 & 0.493 & 0.392 & 0.004 & 0.051 & 0.358 \\
DINOv3-B (Mean/Uniform)  & 0.298 & 0.493 & 0.393 & 0.004 & 0.050 & 0.362 \\

\midrule

DINOv3-L (Attention/Focal) & 0.307 & 0.535 & 0.384 & 0.004 & 0.036 & 0.419 \\
DINOv3-L (Attention/Uniform)  & 0.313 & 0.522 & 0.400 & 0.004 & 0.046 & \textbf{0.452} \\
DINOv3-L (Mean/Focal) & 0.300 & 0.501 & 0.389 & 0.004 & 0.034 & 0.399 \\
DINOv3-L (Mean/Uniform)  & 0.303 & 0.503 & 0.401 & 0.004 & 0.049 & 0.363 \\

\midrule

ResNet50 (Attention/Focal) & 0.268 & 0.449 & 0.348 & 0.004 & 0.030 & 0.353 \\
ResNet50 (Attention/Uniform)  & 0.268 & 0.441 & 0.353 & 0.004 & 0.029 & 0.360 \\
ResNet50 (Mean/Focal) & 0.264 & 0.451 & 0.346 & 0.004 & 0.029 & 0.312 \\
ResNet50 (Mean/Uniform)  & 0.270 & 0.436 & 0.350 & 0.004 & 0.098 & 0.335 \\

\midrule

VideoMAE (Attention/Focal) & 0.301 & 0.505 & 0.382 & 0.004 & 0.052 & 0.424 \\
VideoMAE (Attention/Uniform)  & 0.304 & 0.516 & 0.385 & 0.005 & 0.077 & 0.388 \\
VideoMAE (Mean/Focal) & 0.299 & 0.507 & 0.386 & 0.005 & 0.065 & 0.358 \\
VideoMAE (Mean/Uniform)  & 0.305 & 0.482 & 0.392 & 0.005 & \textbf{0.114} & 0.426 \\

\midrule

VideoMAE-L (Attention/Focal)  & 0.308 & 0.507 & 0.402 & \textbf{0.006} & 0.061 & 0.389 \\
VideoMAE-L (Attention/Uniform)   & 0.307 & 0.492 & 0.414 & 0.005 & 0.043 & 0.385 \\
VideoMAE-L (Mean/Focal)  & 0.318 & 0.515 & 0.430 & 0.005 & 0.048 & 0.365 \\
VideoMAE-L (Mean/Uniform)   & 0.307 & 0.501 & 0.414 & 0.006 & 0.060 & 0.346 \\

\midrule

V-JEPA2 (Attention/Focal)  & 0.298 & 0.471 & 0.400 & 0.004 & 0.029 & 0.421 \\
V-JEPA2 (Attention/Uniform)   & 0.293 & 0.485 & 0.387 & 0.004 & 0.036 & 0.377 \\
V-JEPA2 (Mean/Focal)  & 0.335 & \textbf{0.552} & 0.444 & 0.005 & 0.032 & 0.431 \\
V-JEPA2 (Mean/Uniform)   & \textbf{0.348} & 0.544 & \textbf{0.478} & 0.005 & 0.061 & 0.430 \\
V-JEPA2 (V-Attn/Focal)  & 0.304 & 0.484 & 0.408 & 0.004 & 0.034 & 0.412 \\
V-JEPA2 (V-Attn/Uniform)   & 0.313 & 0.503 & 0.421 & 0.004 & 0.030 & 0.415 \\

\bottomrule
\end{tabular}

\caption{\scriptsize Mean Average Precision (mAP) comparison of backbones and strategies on the \textbf{FishFollow} dataset.}
\label{tab:fishfollow_map_results}
\vspace{-2em}
\end{table*}

\section{Accesibility}
\label{supp:datasheet}
We follow the datasheet proposed in~\cite{gebru2021datasheets} for documenting our WildFin dataset. 

WildFin is available from our project website: \url{https://team-wildfin.github.io/}. The dataset is openly available without categorical access restrictions, under an open licences that permits research use with appropriate citation of the WildFin dataset and paper. 

To support reproducibility and community reuse, we release all code, training configurations, data splits, hyperparameters, and supporting scripts used in our experiments. Comprehensive documentation of data formats, annotation protocols, and metadata schemas are provided in the main paper and in the repository, enabling the research community to reproduce our results and extend the dataset and benchmarks.

\end{document}